\documentclass[letterpaper]{article} 
\usepackage{aaai2026}  
\usepackage{times}  
\usepackage{helvet}  
\usepackage{courier}  
\usepackage[hyphens]{url}  
\usepackage{graphicx} 
\usepackage{natbib}  
\usepackage{caption} 
\usepackage{algorithm}
\usepackage{algorithmic}

\usepackage{graphicx}       
\usepackage{amsmath}        
\usepackage{booktabs}       
\usepackage{verbatim}       
\usepackage{subcaption}     
\usepackage{amsfonts}       

\usepackage{newfloat}
\usepackage{listings}
\DeclareCaptionStyle{ruled}{labelfont=normalfont,labelsep=colon,strut=off} 
\floatstyle{ruled}
\newfloat{listing}{tb}{lst}{}
\floatname{listing}{Listing}
\title{MENDR: Manifold Explainable Neural Data Representations}

\author{
    Matthew Chen\textsuperscript{\rm 1},
    Micky Nnamdi\textsuperscript{\rm 1},
    Justin Shao\textsuperscript{\rm 2},
    Andrew Hornback\textsuperscript{\rm 1},
    Hongyun Huang\textsuperscript{\rm 1},
    Ben Tamo\textsuperscript{\rm 1},
    Yishan Zhong\textsuperscript{\rm 1},
    Benoit Marteau\textsuperscript{\rm 1},
    Wenqi Shi\textsuperscript{\rm 1,3},
    May Dongmei Wang\textsuperscript{\rm 1}
}
\affiliations{
    \textsuperscript{\rm 1}Georgia Institute of Technology, Atlanta, GA 30322 USA\\
    \textsuperscript{\rm 2}Carnegie Mellon University, Pittsburgh, PA 15213\\
    \textsuperscript{\rm 3}UT Southwestern Medical Center, Dallas, TX 75390\\
    
    mchen439@gatech.edu, mnnamdi3@gatech.edu, jshao2@andrew.cmu.edu, ahornback6@gatech.edu, hhuang466@gatech.edu, jtamo3@gatech.edu, yzhong307@gatech.edu, bmarteau3@gatech.edu, wenqi.shi @utsouthwestern.edu, maywang@gatech.edu
}

\begin{document}

\maketitle

\begin{abstract}

Foundation models for electroencephalography (EEG) signals have recently demonstrated success in learning generalized representations of EEGs, outperforming specialized models in various downstream tasks. However, many of these models lack transparency in their pretraining dynamics and offer limited insight into how well EEG information is preserved within their embeddings. For successful clinical integration, EEG foundation models must ensure transparency in pretraining, downstream fine-tuning, and the interpretability of learned representations. Current approaches primarily operate in the temporal domain, overlooking advancements in digital signal processing that enable extracting deterministic and traceable features, such as wavelet-based representations. We propose MENDR (Manifold Explainable Neural Data Representations), a filter bank-based EEG foundation model built on a novel Riemannian Manifold Transformer architecture to resolve these issues. MENDR learns symmetric positive definite matrix embeddings of EEG signals and is pretrained on a large corpus comprising over 4,000 hours of EEG data, decomposed via discrete wavelet packet transforms into multi-resolution coefficients. MENDR significantly enhances interpretability by visualizing symmetric positive definite embeddings as geometric ellipsoids and supports accurate reconstruction of EEG signals from learned embeddings. Evaluations across multiple clinical EEG tasks demonstrate that MENDR achieves near state-of-the-art performance with substantially fewer parameters, underscoring its potential for efficient, interpretable, and clinically applicable EEG analysis. 

\end{abstract}


\section{Introduction}
\label{sec:intro}
Electroencephalography (EEG) is the process of non-invasively measuring the brain’s electric fields \cite{BIASIUCCI2019R80}. Although more advanced brain imaging techniques exist, EEG remains the state-of-the-art paraclinical tool for seizure evaluation. Other medical applications of EEG include assessing patients in comas in intensive care units and evaluating encephalopathies \cite{Rayi2025-ah}, sleep stage classification \cite{e18090272}, emotion recognition \cite{liu2021comparing}, motor imagery classification \cite{altaheri2023deep}, and gait prediction \cite{MOBI}.

Unlike earlier deep learning EEG studies, which only leverage CNN or LSTM modules for specific downstream tasks \cite{EEGNet, LSTM-EEG}, attention-based EEG foundation models have recently shown promise in learning generalized EEG representations for a multitude of downstream EEG tasks \cite{BIOT, LABRAM, CBRAMOD, EEGPT}. Since EEG signals can be formulated as a matrix of real numbers $X \in \mathbb{R}^{C \times T}$ where $C$ is the number of EEG electrodes and $T$ represents the number of sampled time steps, one can adopt recent advances in learning generalized representations of time-series \cite{PatchTST} and scalable patch embeddings \cite{VIT, VIT2} to the EEG domain. 

Although several works have made significant strides in the field of EEG foundation models, many challenges remain: \textbf{Neglecting the Neurophysiological Basis of EEG Signals:} Current EEG foundation models approach EEG as a multivariate time series and primarily leverage deep learning methods applicable to any time series, neglecting neuroscientific principles that could drastically improve the efficiency of EEG foundation models for no cost. For example, models like BIOT, LaBraM, CBraMod, and EEGPT \cite{BIOT, LABRAM, CBRAMOD, EEGPT} all design their architectures to handle a variety of EEG electrode layouts rather than fixing a specific number of electrodes. In EEG analysis, the concept of spline interpolation for reconstructing missing channels has existed since 1987 \cite{SplineInterpolation} and has successfully been applied in harmonizing EEG in the largest-scale EEG study of dementia \cite{PRADO202224, Moguilner2024}. Unlike other multivariate time series domains where specific channels do not exist, EEG is a domain where specific channels \textit{must exist} (unless in extreme cases where a significant chunk of a subject’s brain is missing) but are not recorded, but can be reconstructed to a high degree of accuracy using spline interpolation \cite{Prado2023}. \textbf{Lack of Explainability of Learned EEG Embeddings:} A key barrier to adopting EEG foundation models in healthcare is establishing trust in the decisions of these models \cite{Petersson2022, SujathaRavindran2023}. Moreover, EEG foundation models could discover phenomena in the EEG that are still unknown in neuroscience, and their findings are more interpretable when designing the foundation model’s architecture around explainability. Instead of primarily focusing on teaching an EEG foundation model to generate generalized EEG representations, we can reverse the roles in an explainable EEG foundation model such that the model teaches us why an EEG segment is represented in a particular manner. \textbf{Resource Intensive Foundation Models:} The state-of-the-art EEG foundation model training paradigm is to pretrain transformer models with millions of parameters \cite{BIOT, LABRAM, CBRAMOD, EEGPT, zhang2023brant, FoME}. Although works like BIOT \cite{BIOT} reduce the computational complexity of calculating the attention between patch embeddings by introducing linear transformers, the number of attention calculations still scales linearly with the number of attention modules in the neural transformers and input sequence length at the minimum. Thus, a simple way to further improve the performance-to-efficiency ratio of EEG foundation models is to reduce the number of parameters in these models. In doing so, the models are more robust in generating generalized EEG representations and require less computational power, which is crucial for real-time clinical applications where speed is critical. 
   
We propose a novel EEG foundation model architecture, MENDR (Manifold Explainable Neural Data Representations) to address the aforementioned issues. Inspired by recent applications of wavelet theory to transformers such as Wave-VIT and Wavelet2Vec \cite{wavevit2022, wavelet2vec}, we instead process the wavelet decompositions of the EEG signal rather than the raw EEG signal. Next, we generate patch embeddings of the wavelet decompositions and pass these embeddings into a GNN-based spatial harmonizer to learn a deep spatial interpolation of the EEG signals, ensuring that the electrode layout is fixed to the international 10-20 system. Finally, we pass the spatially harmonized embeddings into an SPD manifold transformer based on the MAtt attention proposed by \cite{pan2022matt}. We want to maximize the explainability of our foundation model while achieving results comparable to those of existing state-of-the-art foundation models. Our contributions are as follows:

\begin{itemize}
    \item \textbf{MENDR: The First Riemannian EEG Foundation Model.} 
    We introduce \textbf{MENDR} (\textit{Manifold Explainable Neural Data Representations}), the first Riemannian EEG foundation model---to our knowledge---to learn the Riemannian manifold structure of EEG covariance features. Pretrained on the \textbf{Temple University Hospital EEG Corpus (TUEG)}, MENDR learns generalized, denoised wavelet and full-signal embeddings using the \textit{Manifold Attention (MAtt)} mechanism proposed by ~\cite{pan2022matt}.
    \item \textbf{GNN-based Autoencoder for Spatial Harmonization.} The GNN-based autoencoder learns a deep spline interpolation of EEG electrode layouts by leveraging known electrode positions and latent time-frequency embeddings of each wavelet patch to enforce a universal EEG layout for any EEG dataset. GNNs have been used in EEG foundation models before \cite{EEGPT}, but not for generating generalized representations.
    \item \textbf{Multi-Resolution Manifold Transformer with Dual-Task Self-Supervised Learning.}
    We design a novel \textit{Manifold transformer architecture} to capture hierarchical wavelet-based SPD embeddings. The model is trained using a dual-task self-supervised objective, combining a \textit{leave-one-out (LOO) contrastive learning task} ~\cite{thapa2024sleepfm} and a \textit{patch-based masked autoencoder reconstruction task}~\cite{VIT}.

    \item \textbf{Explainable and Compressible Representations.}
    We ensure both \textbf{explainability and signal compressibility} by introducing a \textit{reconstructive decoder head}, which enables the accurate recovery of wavelet coefficients from latent embeddings and enhances the interpretability of pre-contextualized signal representations.

    \item \textbf{Geometry-Aware Generalization and Visualization.}
    MENDR exploits the \textbf{Riemannian geometry} of SPD matrices for \textit{ellipsoid-based embedding visualization}, and uses \textbf{graph neural networks} for \textit{geometric harmonization across EEG electrode layouts}, improving generalization and interpretability ~\cite{GEFM}.
\end{itemize}

\section{Methodology}
\label{sec:Methodology}

\begin{figure*}
    \centering
	\includegraphics[width=\linewidth]{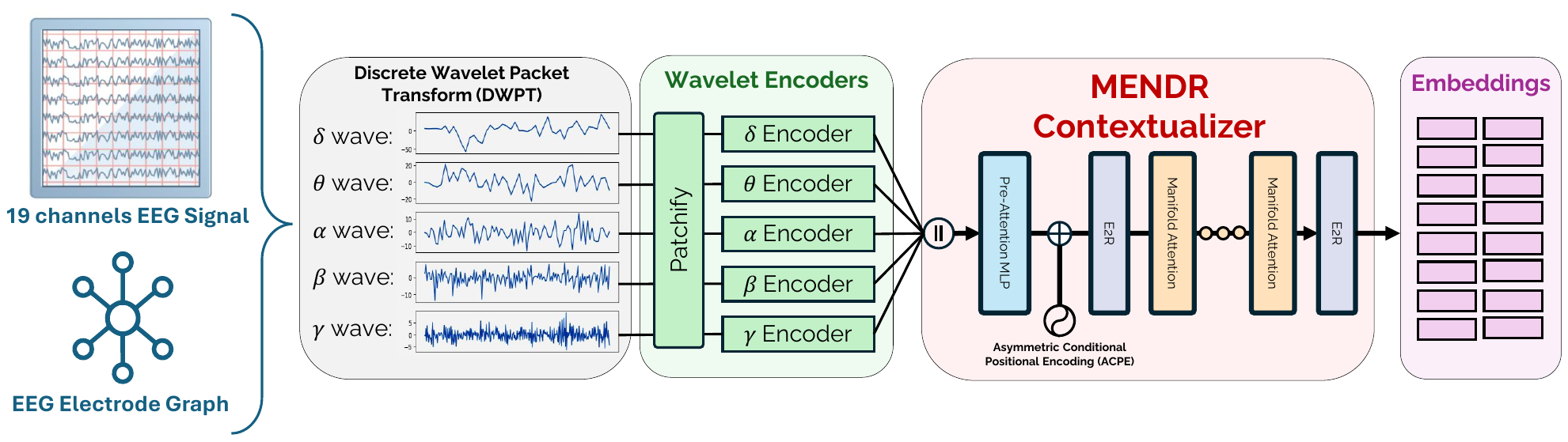}
	\caption[MENDR Overview]{The MENDR pretraining process. EEG signals are first decomposed into the $\delta$, $\theta$, $\alpha$, $\beta$, $\gamma$ frequency decompositions and then fed into their respective encoders to generate MENDR wavelet features. Afterwards, the features are fed into a reconstruction decoder and a contextualizer. MENDR performs Riemannian self-supervised learning with one or two contrastive loss tasks depending on the contextualizer size.}
	\label{fig:MENDRPretrainingOverview}
\end{figure*}

In this section, we present the architecture of MENDR (Figure \ref{fig:MENDRPretrainingOverview}). MENDR builds on concepts from BENDR~\cite{BENDR} and Wav2Vec~\cite{wav2vec}, combining ideas from both foundational models. The model comprises an encoder that extracts multi-scale wavelet features from the EEG signal patches, followed by a contextualizer that models temporal dependencies across patches.

\textbf{MENDR Autoencoder:} For each wavelet frequency band, each patch is \textbf{featurized into a two-second SPD matrix embedding} in contrast to one-second patches traditionally converted into a sequence of vector embeddings. More information about preprocessing the EEG data into patches can be found in Section~\ref{subsec:DataPreprocess} of the Appendix. Each frequency band has its corresponding patch length $T_{band}$, equivalent to two seconds of time series points within that frequency range. MENDR then employs an encoder–decoder architecture that learns a compact representation of the corresponding band-limited signal and reconstructs the original wavelet patches from that embedding. Each encoder begins with three GNN transformer blocks, each comprising a GATConv \cite{GAT} attention module, two linear layers, and layer norms in between. We refer to these three blocks as the GNN Spatial Harmonizer. During pretraining, we employ channel dropout before feeding the wavelet decompositions, allowing the GNN Spatial Harmonizer to learn to impute missing channels and perform spline interpolation based on the known/missing electrode locations from the electrode graph. Then, the outputs are fed into a Squeeze-and-Excitation block \cite{hu2018squeeze} to further capture the relationships between different channels. Finally, the harmonized embeddings, which now have 19 channels, are fed into three convolution blocks, each consisting of a 2D convolution, group normalization, and GELU activation. This produces a sequence of patch embeddings with length $t^{\text{band}}_{\text{hidden}}$. Unlike prior works that rely on Short-Time Fourier Transform (STFT) features, MENDR extracts overlapping sub-patches directly from the wavelet-transformed signal using learnable convolutions with no overlap. A visualization of the architecture is presented in Appendix Figure \ref{fig:MENDREncoder}. 

\begin{figure*}
    \centering
	\includegraphics[width=\linewidth]{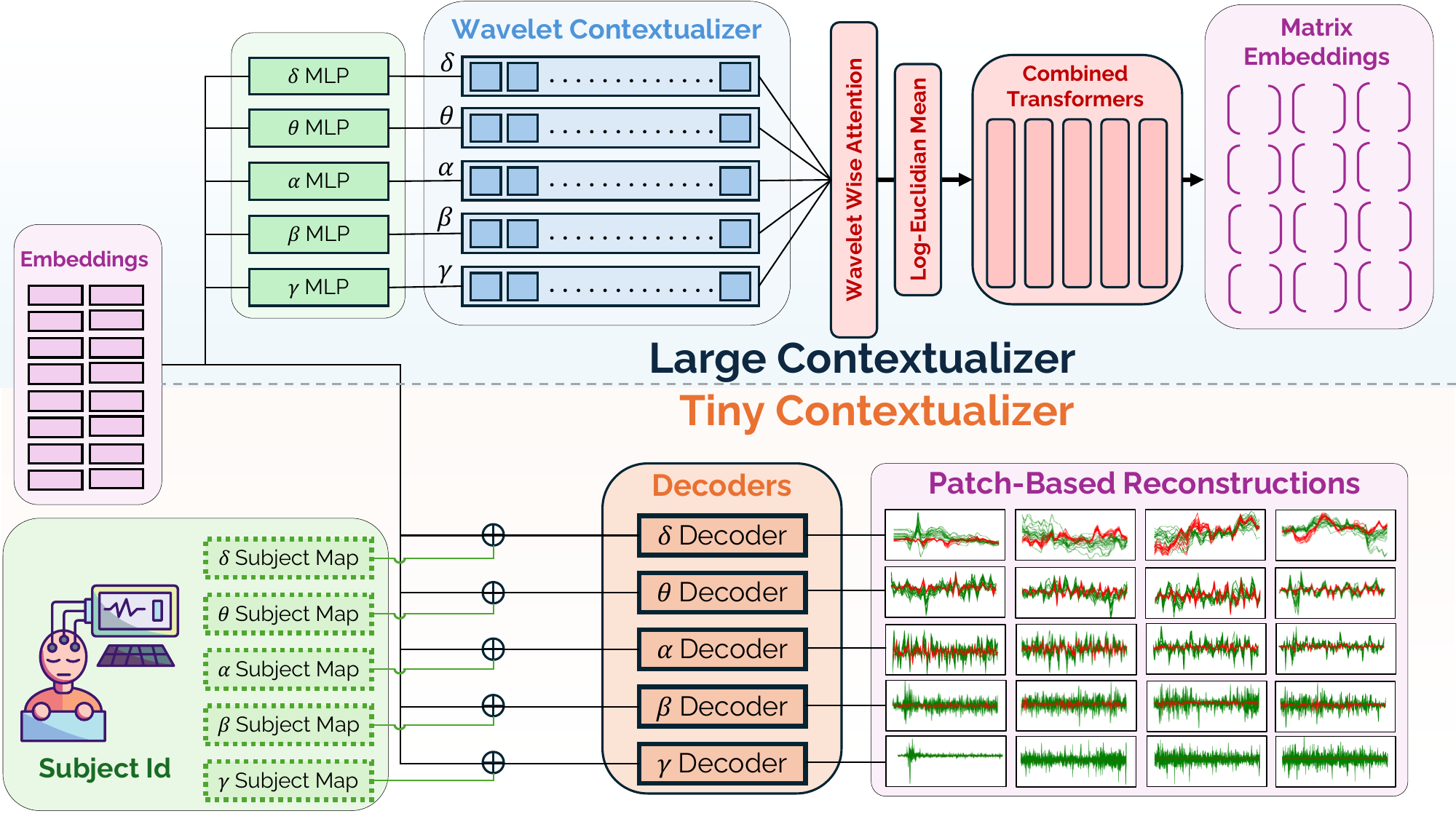}
	\caption[MENDR Contextualizers]{The MENDR Tiny and Large Contextualizer architectures and the “Decoder” part of the Autoencoder, which is only used during pre-training. Note that the tiny contextualizer is identical to the large contextualizer, except without the wavelet contextualizer layers. The autoencoder encoder layers are trained in conjunction with the decoder layers to reconstruct the original wavelet patches, ensuring that the wavelet embeddings retain physical interpretability.}
	\label{fig:MENDRContextualizers}
\end{figure*}

\textbf{MENDR Tiny/Large Contextualizer.} We feed the wavelet patch features into a tiny or a large contextualizer. Let $n$ represent the total number of patches. Both contextualizers begin with a two-layer multi-layer perceptron (MLP) layer, which only learn along the $t^{\text{band}}_{\text{hidden}}$ dimension. Then, in the tiny contextualizer and the wavelet contextualizer of the large contextualizer, Euclidean positional encodings are generated. Like CBraMod \cite{CBRAMOD}, we employ an asymmetric conditional positional encoding (ACPE) scheme to facilitate the encoding of spatial and temporal information for the contextualizer. Our implementation of the ACPE is devised of a 2D convolutional layer with kernel $(k_{t}, k_{s})$ where $k_{t}$ is the kernel size of the temporal dimension, $k_{s}$ is the kernel size of the spatial/channel dimension and $k_{t} > k_{s}$. We set $k_s$ equal to the number of channels, $19$, and $k_{t} = 3$ so that the kernel can encode short-range temporal position information and long-range spatial positional information. Then, we feed the super patch embeddings into the positional encoder to generate the ACPE $S^{p} = \{ s^{p}_{j}| j \in [1, 2, \ldots, n]\}$, where $S^{P} \in \mathbb{R}^{19 \times n \times t^{\text{band}}_{\text{hidden}}}$ and $s^{p}_{j} \in \mathbb{R}^{19 \times t^{\text{band}}_{\text{hidden}}}$. We then add the ACPE to the patch embeddings: 

\begin{equation}
    S^{o} = S + S^{p} = \{s_{j} + s^{p}_{j}| j \in [1, 2, \ldots, n] \}
\end{equation}

where $S^{o} \in \mathbb{R}^{19 \times n \times t^{\text{band}}_{\text{hidden}}}$ is the set of EEG patch embeddings with the ACPE and $n$ represents the total number of patches in the sample. Then, we project the features onto the SPD manifold by calculating the Sample Covariance Matrix (SCM) for each patch. Afterwards, we perform trace-normalization and add a small $\epsilon = 1 * 10^{-5}$ on each diagonal after normalization, identical to \cite{pan2022matt}, with the caveat that we also add $\epsilon$ before trace-normalization as well for numerical stability and to guarantee a well-defined SPD matrix:

\begin{equation}
      SCM = \frac{SCM}{tr(SCM) + \epsilon \mathbf{I}} + \epsilon \mathbf{I} \in \mathbb{R}^{n \times 19 \times 19} 
    \label{eq:BatchTraceNorm}
\end{equation}

where $\mathbf{I}$ is the identity matrix. If the contextualizer is large, we reduce the dimensionality of each SCM matrix embedding to a smaller square dimension, $6$, using a Bilinear layer as proposed by SPDNet \cite{huang2017riemannian}. We need to reduce the dimensionality of each matrix embedding because the forward and backward passes of a bilinear layer and ReEig non-linear activation function require calculating the Singular Value Decomposition three times in total \cite{huang2017riemannian}, which is an $O(n^{3})$ operation. Then, the SPD embeddings are passed through multiple layers of an Riemannian SPD Manifold analogue of the original Transformer encoder architecture from \cite{vaswani2017attention}. More details on differentiating the SVD, the mathematics of Riemannian manifold optimization, and translating them into a self-supervised learning setting are included in Section \ref{subsec:Mathematical Preliminaries} of the Appendix. The architectures of the tiny and large contextualizers are found in Figure \ref{fig:MENDRContextualizers}.  

\subsection{MENDR Pretraining}

We train MENDR in two phases: autoencoder training and contextualizer training like \cite{LABRAM}. 

\textbf{Autoencoder Pretraining}
The autoencoder is first trained primarily to learn subject-independent embeddings by giving only the decoder information about the subject ID. The motivation for including the subject ID relies on the fact that the EEG subject contributes the most variance to differences in EEG dynamics \cite{melnik2017systems}. Additionally, as noted in \cite{van2017neural}, the authors achieved success in learning subject-agnostic speech recording embeddings by passing the speaker ID only to the decoder. Thus, given encoder output $z_{e}(X)$, the decoding/reconstruction is calculated as $z_{d}(X|\text{Subject ID}) = z_{d}(z_{e}(X) + z_{d}^{\textit{Subject\_ID}})$ where $z_{d}^{\textit{Subject\_ID}}$ is a learnable parameter unique to each subject ID in the pretraining dataset. 

\textbf{Reconstruction Loss.} To ensure that the autoencoder is properly learning robust wavelet embeddings that contain the same information as the original data, the wavelet coefficient time series, we define the wavelet reconstruction loss as follows:

\begin{equation}
 \mathcal{L}^{\text{wavelet}}_{\text{recon}}(X, X_{\text{recon}}) = \|X - X_{\text{recon}}\|^{2}  
\end{equation}

The equation represents the mean squared error (MSE) reconstruction error between the actual and reconstructed signals in the time domain. 

\textbf{Wavelet LOO Loss.} Next, if the contextualizer is large, we break pretraining of the contextualizer into two steps: wavelet LOO pretraining for the wavelet contextualizers and combined masked autoencoding for the "combined" (Log-Euclidean mean) of the wavelet embeddings. Although we deal with a single modality, EEG, we still break the single EEG time series into five different frequency bands, giving us five different resolutions for analyzing a single EEG patch. Thapa et al., while dealing with multiple modalities, proposed the novel Leave-One-Out (LOO) contrastive learning method while training their SleepFM foundation model \cite{thapa2024sleepfm}. LOO aims for a predictive task where one modality tries to identify the corresponding embeddings from the remaining modalities. Consider each frequency band as a “modality” to view LOO for our wavelet bands, as each frequency band identifies embeddings of the other frequency bands. LOO contrastive loss also has a medical neuroscientific basis, as different frequency bands, such as $\delta$ and $\theta$ bands, have been known to couple and have their coupling altered in certain neurological diseases \cite{carracedo2013neocortical, wirt2021altered}. Thus, wavelet LOO contrastive learning has a neuropsychological motivation because it helps the embeddings learn inter-filter-bank relationships that can be mapped to empirical neuroscientific results on band coupling. 

Since our embeddings lie on a Riemannian manifold, we cannot use the Euclidean mean as a representation of our embeddings like \cite{thapa2024sleepfm}, and instead rely again on the Log Euclidean mean. If we let $\mathbf{\bar{A}_{\text{band}}^{\neq \text{band}}}$ be the LEM of the other bands, excluding the band under analysis, our modified LOO loss is:

\begin{equation}
    \mathcal{L}_{\text{band}}^{\text{LOO}} = -\log{\frac{\exp(sim(\mathbf{A}^{\text{band}}, \mathbf{\bar{A}_{\text{band}}^{\neq \text{band}}})) * \exp(\tau)}{\sum_{m=1}^{N_{\text{negatives}}} exp(sim(\mathbf{A}^{\text{band}}, \mathbf{\bar{A}_{\text{band}}^{\neq \text{band}}})) * \exp(\tau)}} 
\end{equation}

where $\delta_{L}$ represents the Log-Euclidean Metric (LEM), $\tau$ represents a learnable temperature scalar parameter, and $N_{\text{negatives}}$ represents the total number of distractors (identical to how it’s defined in BERT \cite{devlin2019bert}). Furthermore, $sim(\mathbf{X}, \mathbf{Y})$ represents the similarity between $\mathbf{X}$ and $\mathbf{Y}$ when $X, Y \in \mathcal{M}_{SPD}$ based on the Log-Euclidean distance. As defined in \cite{pan2022matt}:

\begin{equation}
    sim(\mathbf{X}, \mathbf{Y}) = \frac{1}{1 + \log(1 + \delta_{L}(\mathbf{X}, \mathbf{Y}))}
\end{equation}

Note that $sim(\mathbf{X}, \mathbf{Y})$ is a \textit{strictly decreasing function} of distance $[0, \infty) \mapsto [0, 1]$ and how similarity between a query and key is calculated in MAtt \cite{pan2022matt}.  

\textbf{Combined Masked Autoencoding Loss}  
In both the tiny and large contextualizers, the wavelet embeddings for an EEG patch are combined into their LEM to represent an embedding of the whole patch. To be consistent with these studies while dealing with SPD matrix embeddings, we utilize \textbf{patch-based masked EEG reconstruction} to learn generalized SPD matrix embeddings of EEG. More specifically, our combined signal transformer serves as a reconstruction head, which takes in a temporally or epoch-masked LEM embedding input and attempts to predict the original LEM SPD matrix. Mathematically speaking, each wavelet input $\mathbf{X}$ is broken up into $n$ patches while in SPD manifold space: $\mathbf{X} \mapsto \{\mathbf{A_{1}}, \mathbf{A_{2}}, \mathbf{A_{3}}, \ldots, \mathbf{A_{n}}\}$, each patch representing $2$ seconds. Note that each $\mathbf{A}$ represents the LEM of the wavelet embeddings. Similar to BERT \cite{devlin2019bert}, with probability $\mathbb{P}_{mask} = 0.2 \approx 0.15$ from each input wavelet segment and replace it with a learnable SPD mask $M \in \{P \in \mathbb{R}^{n\times n}| P = P^{T}, v^{T}Pv > 0, \forall v \in \mathbb{R}^{n} - \{0\} \}$ such that $\lfloor{ \mathbb{P}_{mask} \times n} \rfloor$ of the epochs are masked. Say we have $10$ patches. For the $5$ wavelet embeddings at patches $6$ and $9$, we replace them with $M$ and calculate the LEM for each patch, simply using $M$ itself. So we have:
\vspace{-2mm}
\begin{equation}
    \{\mathbf{A_{1}}, \mathbf{A}_{2}, \mathbf{A_{3}}, \mathbf{A_{4}}, \mathbf{A_{5}}, M, \mathbf{A_{7}}, \mathbf{A_{8}}, M, \mathbf{A_{10}}\}
\end{equation}

and then feed it through the combined signal transformer. The transformer outputs a new representation:
\vspace{-2mm}
\begin{equation}
    \{\hat{\mathbf{A_{1}}}, \hat{\mathbf{A_{2}}}, \hat{\mathbf{A}_{3}}, \hat{\mathbf{A_{4}}}, \hat{\mathbf{A_{5}}}, \hat{\mathbf{A_{6}}}, \hat{\mathbf{A_{7}}}, \hat{\mathbf{A_{8}}}, \hat{\mathbf{A_{9}}}, \hat{\mathbf{A_{10}}}\}
\end{equation}

Note that we only calculate the MSE for masked patches. Thus, our loss is defined as:

\begin{equation}
    \mathcal{L}^{MAE}_{\text{combined}} = \|\log(\lambda_{A_{\text{masked}}}) - \log(\lambda_{\hat{A}_{\text{masked}}}) \|^{2}
\end{equation}

where $\log(\lambda_{A_{\text{masked}}})$ and $\log(\lambda_{\hat{A}_{\text{masked}}})$ represent the Euclidean eigenvalues of $A_{\text{masked}}$ and $\hat{A}_{\text{masked}}$. Since $M$ is a learnable SPD matrix mask, we need to ensure that $M$ remains SPD after backpropagation. We could modify the backpropagation, similar to \cite{huang2017riemannian}, and calculate a gradient on the SPD manifold. However, a more straightforward method is to note that any SPD matrix $M$ is equal to the product of some matrix and its transpose through the Cholesky Decomposition. 

\textit{Short Proof}: If $M \in \mathbb{R}^{n \times n}$ is SPD, then $M = LL^{T}$ for some $L \in \mathbb{R}^{n \times n}$. Since $M$ is symmetric, it must be orthogonally diagonalizable: $M = PDP^{T}$ for some diagonal matrix $D$ and orthogonal matrix $P$. By the diagonalization theorem, the diagonal values of $D$ are the eigenvalues of $A$ and are all positive. Since the diagonal entries are all positive, a matrix $F$ must exist such that $F^{2} = D$. Essentially, the diagonal entries are the square roots of the original diagonal matrix. Therefore:

\begin{equation}
\begin{split}
    M &= PF^{2}P^{T} = (PF)(FP^{T})\\
      &= (PF)(F^{T}P^{T}) = (PF)(PF)^{T}
\end{split}
\end{equation}
If we let $L = PF$, we are done. Thus, we indirectly optimize on the SPD manifold by initializing a random $L \in \mathbb{R}^{n \times n}$ and letting $M = LL^{T}$. In other words, we can still use Euclidean gradients to optimize over the SPD manifold by the product rule:

\begin{equation}
    \mathbf{d}M = (\mathbf{d}L)L^{T} + L(\mathbf{d}L^{T})
\end{equation}
\section{Experimental Results}
\label{sec:Results}

\subsection{Pre-training Settings} We implemented and ran experiments for MENDR on Python 3.10.16 on PyTorch 2.5.1 + CUDA 12.8. Since each EEG recording in the TUEG pretraining dataset is one minute, we generate 30 two-second super-patches from each segment. The batch size is set to $256$ in all phases of pretraining, the learning rate for pretraining the autoencoder was set to $1 * 10^{-4}$ and for all contextualizers at $1 * 10^{-3}$, and L2 weight decay is set to $0.001$ in all cases. We use a CosineAnnealingLR due to the complex training dynamics in our model. To ensure the pretraining converges at the last epoch, we set $T_{\text{max}}$ always equal to the number of training epochs. We used 30 epochs for pre-training the autoencoder, five epochs for training the combined contextualizers in both models, and five epochs for pre-training the wavelet. Note that in each subsequent phase of training, we use the weights from the previous module. Pretraining was performed on a single machine equipped with an Intel Xeon Platinum 8562Y+ CPU, 16 allocated cores, 512 GB of RAM, and a single NVIDIA HGX H200 GPU with approximately 140 GB of VRAM. However, pretraining can be performed on GPUs with less VRAM, as not all of the VRAM is utilized during this process. One can find specific hyperparameter details in Appendix Table \ref{table:PretrainHyperparameters}.

Note that we also include both the tiny and large models with “high” frequencies (64-128 Hz), i.e., models pre-trained and then trained on the TUAB/TUEV. The addition of the “high” frequencies requires another autoencoder/contextualizer for the specific frequency range. Note that this refers to the “D” block in Appendix Figure \ref{fig:WaveletPacketDecomposition}.
\subsection{Downstream Task Evaluation and Comparison with State-of-the-Art} 
To evaluate whether MENDR generates generalized EEG representations, we assess its performance on five downstream tasks, three of which are presented in the main paper and two in the appendix. First, we present the results for the Temple University Abnormal EEG Corpus (TUAB) by \cite{Lopez2015} and the Temple University EEG Events Corpus (TUEV) from \cite{obeid2016temple}. Note that both TUAB and TUEG are subsets of the TUEG pretraining dataset, but we did not include task labels during pretraining. We emphasize these two datasets because most previous EEG foundation model studies use them for comparing other EEG foundation models \cite{BIOT, LABRAM, CBRAMOD, EEGPT}. Specific hyperparameter values can be found in the appendix.
\paragraph{Metrics} Identical to \cite{BIOT, LABRAM, CBRAMOD, EEGPT}, we use \textbf{Balanced Accuracy}, which is the average recall for each class, for the TUAB and TUEV datasets. For TUAB (binary classification), we also calculate the area under the precision-recall curve (\textbf{AUC-PR}) and the area under the receiver operating characteristic curve (\textbf{AUROC}). For TUEV (multi-class classification), we use \textbf{Cohen’s Kappa}, a metric that measures the agreement between the model’s predictions and the actual labels, and the \textbf{Weighted-F1} score, which is the harmonic mean of the precision and recall. 

\textbf{Comparison} We include the results of the models from \cite{BIOT}, which consists of the BIOT foundation model and specialist models, and other recent state-of-the-art foundation models \cite{LABRAM, CBRAMOD, EEGPT}. Unless otherwise stated, the results from these baseline models are from existing studies. For the TUAB dataset, instead of classifying one-second EEG patches as normal/abnormal, we classify super-patches as normal or abnormal. Additionally, for the TUEV dataset, since each event is 1 second long, and all baselines preprocess the EEG into five-second samples with the event at the center, we preprocess the data into 6-second samples. Like other studies, we run downstreaming $5$ times with different random seeds and calculate the average and standard deviation of the metric across the $5$ trials. Unlike other studies, we do not further split the training dataset into training and validation. Therefore, instead of a simple MLP decoder, we use a two-layer transformer to predict the sleep stage of each sequence of patches in the context of the other sequence of patches. 

\begin{table*}[ht]
    \centering
    \caption{Results of Different Methods on TUAB and TUEV datasets}
    \label{tbl:TUABTUEVResultTable}
    \resizebox{\linewidth}{!}{%
    \begin{tabular}{cccccccc}
        \hline\hline
        & & \multicolumn{3}{c}{TUAB (Binary Classification)} & \multicolumn{3}{c}{TUEV (6-class Classification)} \\
        \cmidrule(r){3-5} \cmidrule(r){6-8}
        Methods & Model Size & Balanced Accuracy & AUC-PR & AUROC & Balanced Accuracy & Cohen’s Kappa & Weighted-F1 \\
        \hline
        \textbf{MENDR-Tiny ($\delta, \theta, \alpha, \beta, \gamma$)} & 1.815M & $0.7786 \pm 0.0009$ & $0.8678 \pm 0.0005$ & $0.8667 \pm 0.0005$ & $0.5446 \pm 0.0113$ & $0.5046 \pm 0.0081$ & $0.7365 \pm 0.0054$\\
        \textbf{MENDR-Tiny ($\delta, \theta, \alpha, \beta, \gamma, \text{high}$)} & 6.996M & $0.7796 \pm 0.0025$ & $0.8731 \pm 0.0018$ & $0.8739 \pm 0.0019$ & $0.5294 \pm 0.0138$ & $0.4704 \pm 0.0329$ & $0.7205 \pm 0.0220$\\
        \textbf{MENDR-Large ($\delta, \theta, \alpha, \beta, \gamma$)} & 1.817M & $0.7838 \pm 0.0048$ & $0.8672 \pm 0.0055$ & $0.8658 \pm 0.0058$ & $0.4686 \pm 0.0276$ & $0.2742 \pm 0.0330$ & $0.5898 \pm 0.0243$\\
        \textbf{MENDR-Large ($\delta, \theta, \alpha, \beta, \gamma, \text{high}$)} & 6.500M & $0.7981 \pm 0.0059$ & $0.8811 \pm 0.0038$ & $0.8791 \pm 0.0031$ & $0.4778 \pm 0.0219$ & $0.3499 \pm 0.0457$ & $0.6491 \pm 0.0293$\\
        \hline
        SPaRCNet \cite{Jing2023-vp} & 0.79M & $0.7896 \pm 0.0018$ & $0.8414 \pm 0.0018$ & $0.8676 \pm 0.0012$ & $0.4161 \pm 0.0262$ & $0.4233 \pm 0.0181$ & $0.7024 \pm 0.0104$ \\
        ContraWR \cite{info:doi/10.2196/46769} & 1.6M & $0.7746 \pm 0.0041$ & $0.8421 \pm 0.0104$ & $0.8456 \pm 0.0074$ & $0.4384 \pm 0.0349$ & $0.3912 \pm 0.0237$ & $0.6893 \pm 0.0136$ \\
        CNN-Transformer \cite{peh2022transformer} & 3.2M & $0.7777 \pm 0.0022$ & $0.8433 \pm 0.0039$ & 0.8461 $\pm 0.0013$ & $0.4087 \pm 0.0161$ & $0.3815 \pm 0.0134$ & $0.6854 \pm 0.0293$ \\
        FFCL \cite{li2022motor} & 2.4M & $0.7848 \pm 0.0038$ & $0.8448 \pm 0.0065$ & $0.8569 \pm 0.0051$ & $0.3979 \pm 0.0104$ & $0.3732 \pm 0.0188$ & $0.6783 \pm 0.0120$\\
        ST-Transformer \cite{song2021transformer} & 3.5M & $0.7966 \pm 0.0023$ & $0.8521 \pm 0.0026$ & $0.8707 \pm 0.0019$ & $0.3984 \pm 0.0228$ & $0.3765 \pm 0.0306$ & $0.6823 \pm 0.0190$ \\
        BIOT \cite{BIOT} & 3.2M & $0.7959 \pm 0.0057$ & $0.8792 \pm 0.0023$ & $0.8815 \pm 0.0043$ & $0.5281 \pm 0.0225$ & $0.5273 \pm 0.0249$ & $0.7492 \pm 0.0082$ \\
        LaBraM-Base \cite{LABRAM} & 5.8M & $0.8140 \pm 0.0019$ & $0.8965 \pm 0.0016$ & $0.9022 \pm 0.0009$ & $0.6409 \pm 0.0065$ & $0.6637 \pm 0.0093$ & $0.8312 \pm 0.0052$  \\
        LaBraM-Large \cite{LABRAM} & 46M & $0.8226 \pm 0.0015$ & $0.9130 \pm 0.0005$ & $0.9127 \pm 0.0005$ & $0.6581 \pm 0.0156$ & $0.6622 \pm 0.0136$ & $0.8315 \pm 0.0040$ \\
        LaBraM-Huge \cite{LABRAM} & 369M & $0.8258 \pm 0.0011$ & $0.9204 \pm 0.0011$ & $0.9162 \pm 0.0016$ & $0.6616 \pm 0.0170$ & $0.6745 \pm 0.0195$ & $0.8329 \pm  0.0086$ \\
        CBraMod \cite{CBRAMOD} & 4.0M & $0.8289 \pm 0.0022$ & $0.9258 \pm 0.0008$ & $0.9227 \pm 0.0011$ & $0.6671 \pm 0.0107$ & $0.6772 \pm 0.0096$ & $0.8342 \pm 0.0064$ \\
        EEGPT-Tiny \cite{EEGPT} & 4.7M & $0.7959 \pm 0.0021$ & - & $0.8716 \pm 0.0041$ & $0.5670 \pm 0.0066$ & $0.5085 \pm 0.0173$ & $0.7535 \pm 0.0097$\\
        EEGPT-Large \cite{EEGPT} & 25M & $0.7983 \pm 0.0030$ & - & $0.8718 \pm 0.0040$ & $0.6232 \pm 0.0114$ & $0.6351 \pm 0.0134$ & $0.8187 \pm 0.0063$ \\
        \hline
        \hline
    \end{tabular}%
    }
\end{table*}

Although we distinguish the sizes of the MENDR models in our naming, the larger MENDR model is only slightly larger than the tiny model compared to the size of existing EEG foundation models because we use a lower dimensionality, but more layers, for learning the manifold in the larger contextualizer. Furthermore, when including the “high” frequencies, the number of parameters in the autoencoder scales up faster than in the contextualizer, resulting in the Tiny model with the high frequencies having slightly more parameters than its Large counterpart. Noticeably, MENDR outperforms several existing EEG foundation models on the TUAB dataset and is comparable to BIOT \cite{BIOT} for the TUEV dataset. We also perform label smoothing of $0.1$, as in other works, for the CrossEntropyLoss of the TUEV dataset. The performance-to-parameter ratio of MENDR indicates that scaling the parameter size of EEG foundation models is not the most efficient method of generating better generalized EEG representations. 

\subsection{Scaling Data Size and Scaling Parameter Size}
\begin{figure}[b!]
    \centering 
    \resizebox{\linewidth}{!}{%
\begin{subfigure}{0.33\textwidth}
  \includegraphics[width=\linewidth]{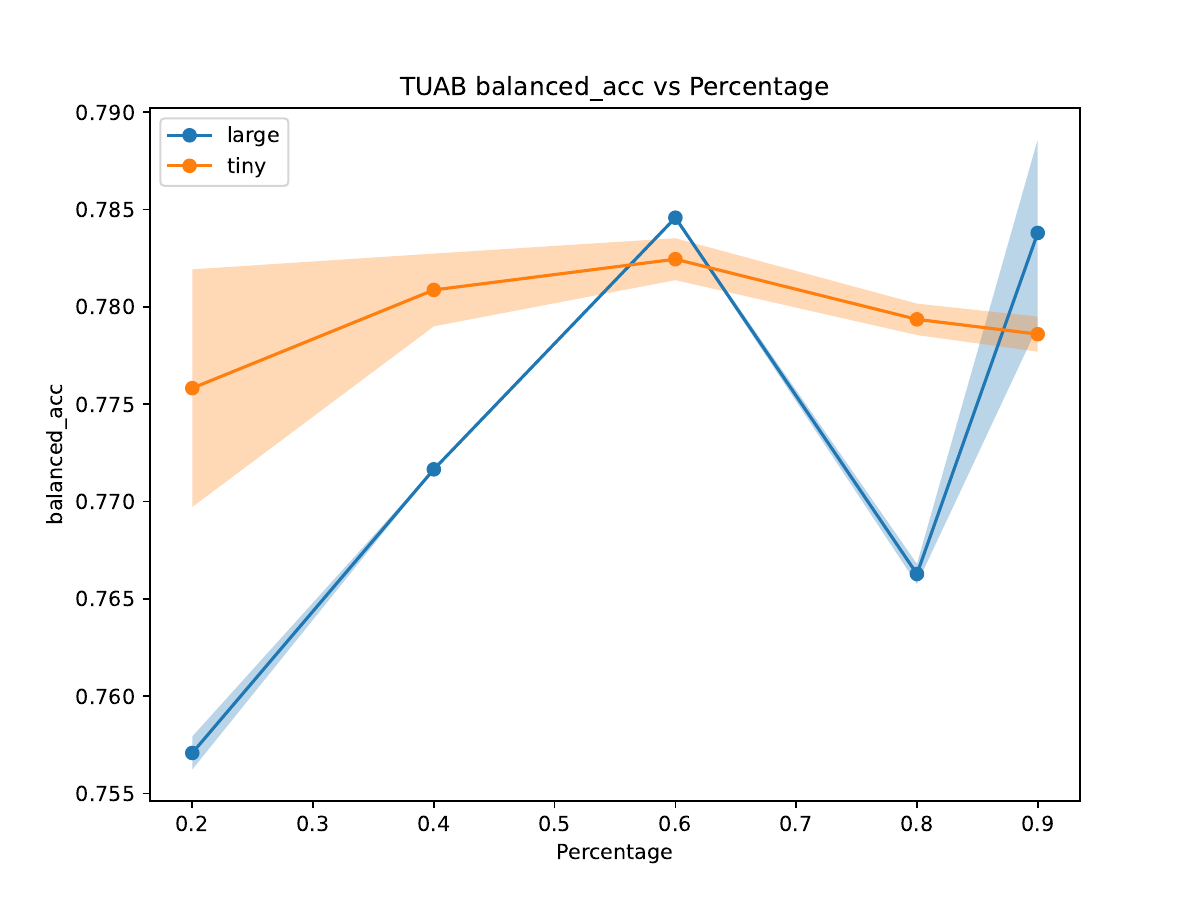}
  \label{fig:1}
\end{subfigure}\hfil 
\begin{subfigure}{0.33\textwidth}
  \includegraphics[width=\linewidth]{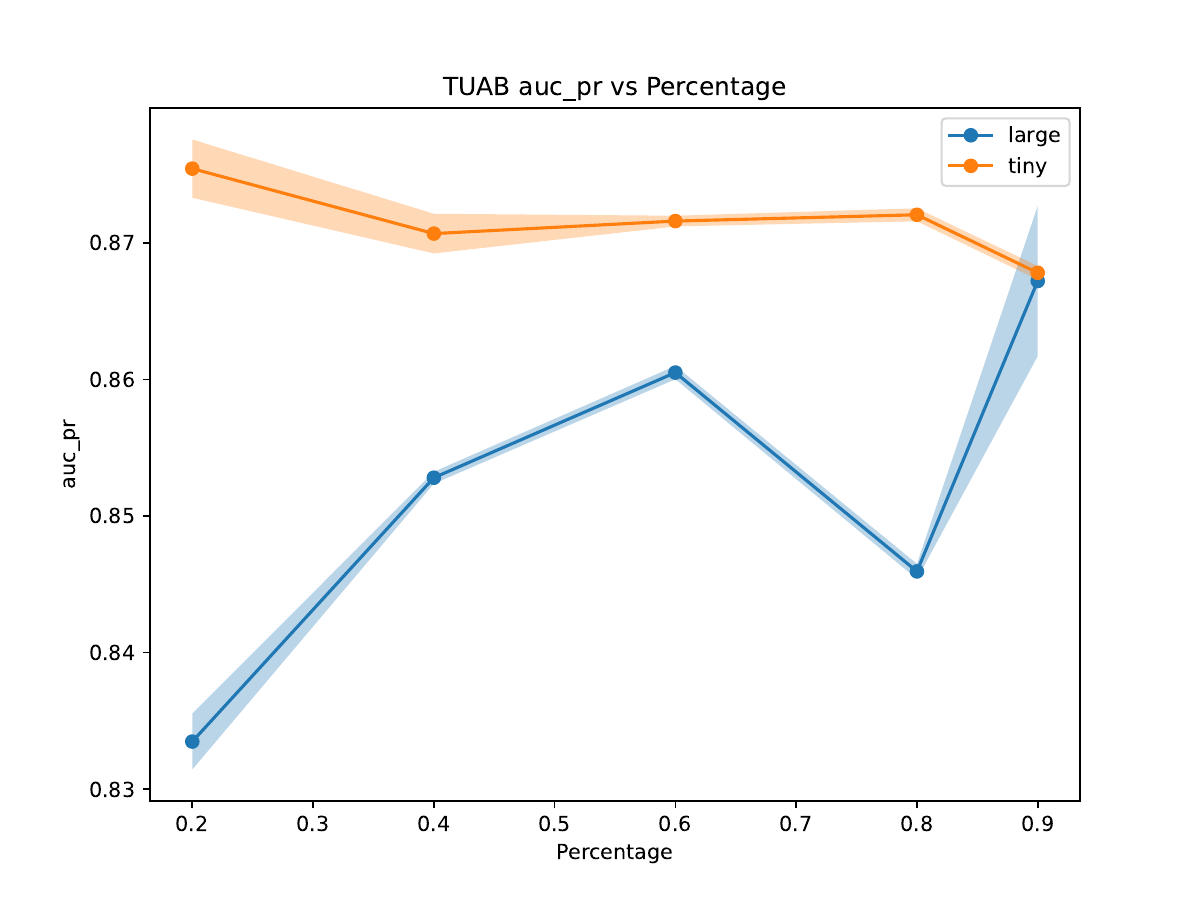}
  \label{fig:2}
\end{subfigure}\hfil 
\begin{subfigure}{0.33\textwidth}
  \includegraphics[width=\linewidth]{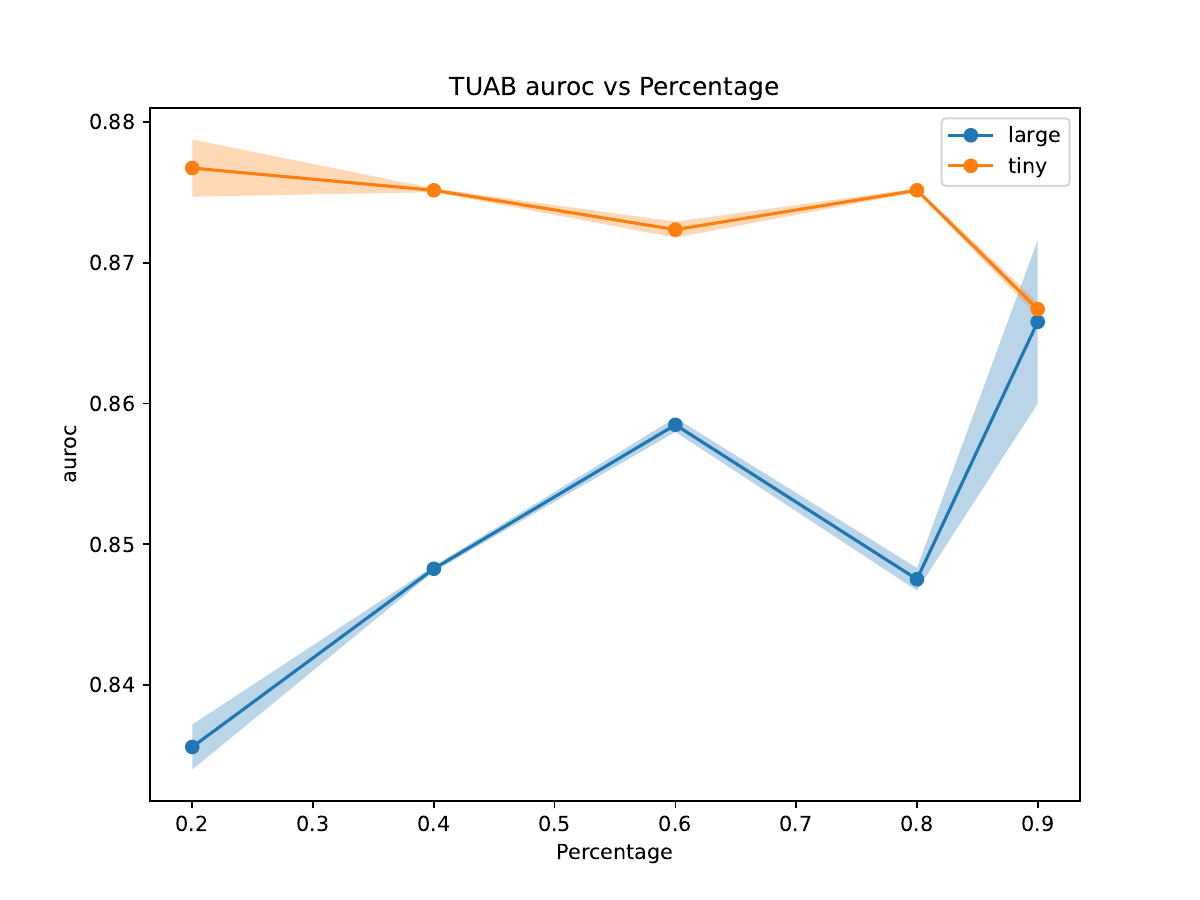}
  \label{fig:3}
\end{subfigure}}
\caption{A performance comparison on TUAB of MENDR Tiny and Large as pretraining data increases.}
\label{fig:scaling_images1}
\end{figure}

\begin{figure}
\centering
 \resizebox{\linewidth}{!}{%
\begin{subfigure}{0.33\textwidth}
  \includegraphics[width=\linewidth]{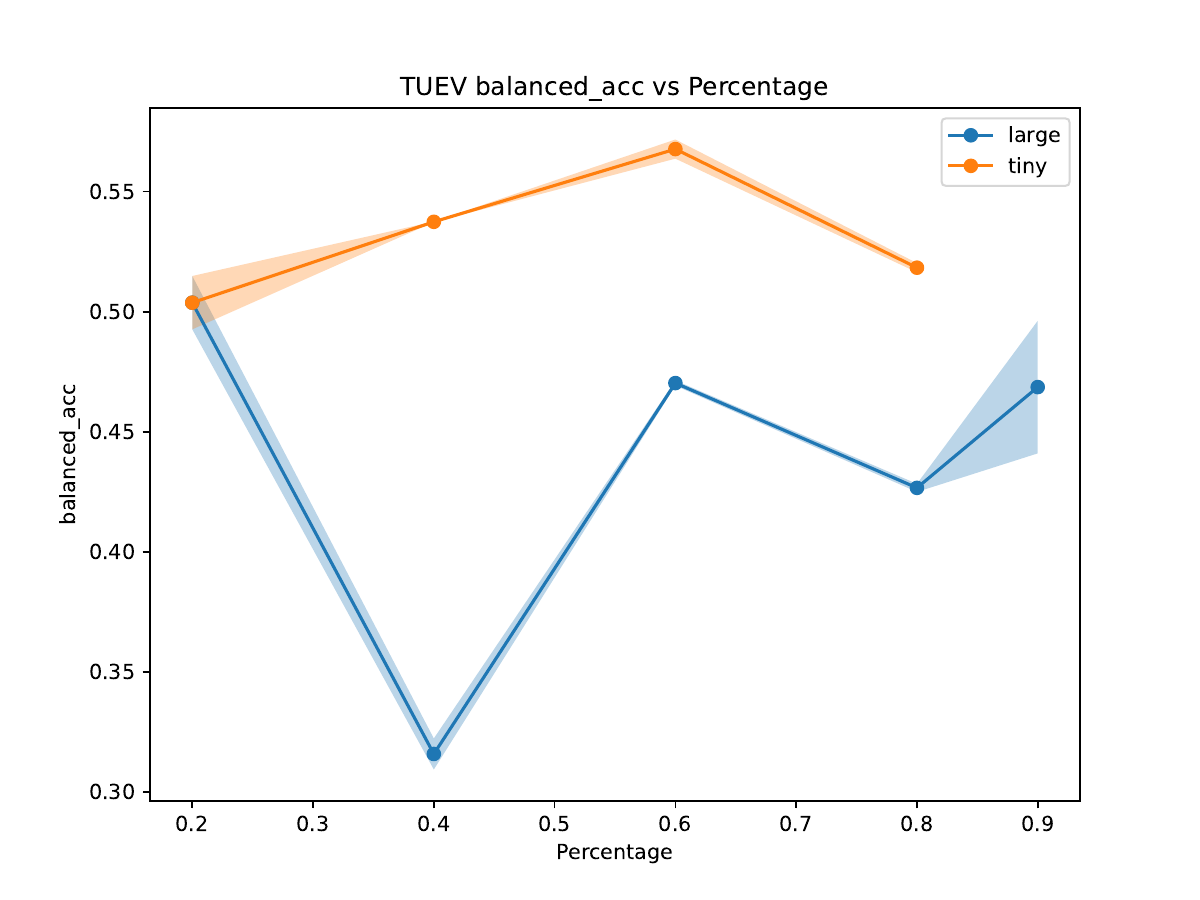}
  \label{fig:4}
\end{subfigure}\hfil 
\begin{subfigure}{0.33\textwidth}
  \includegraphics[width=\linewidth]{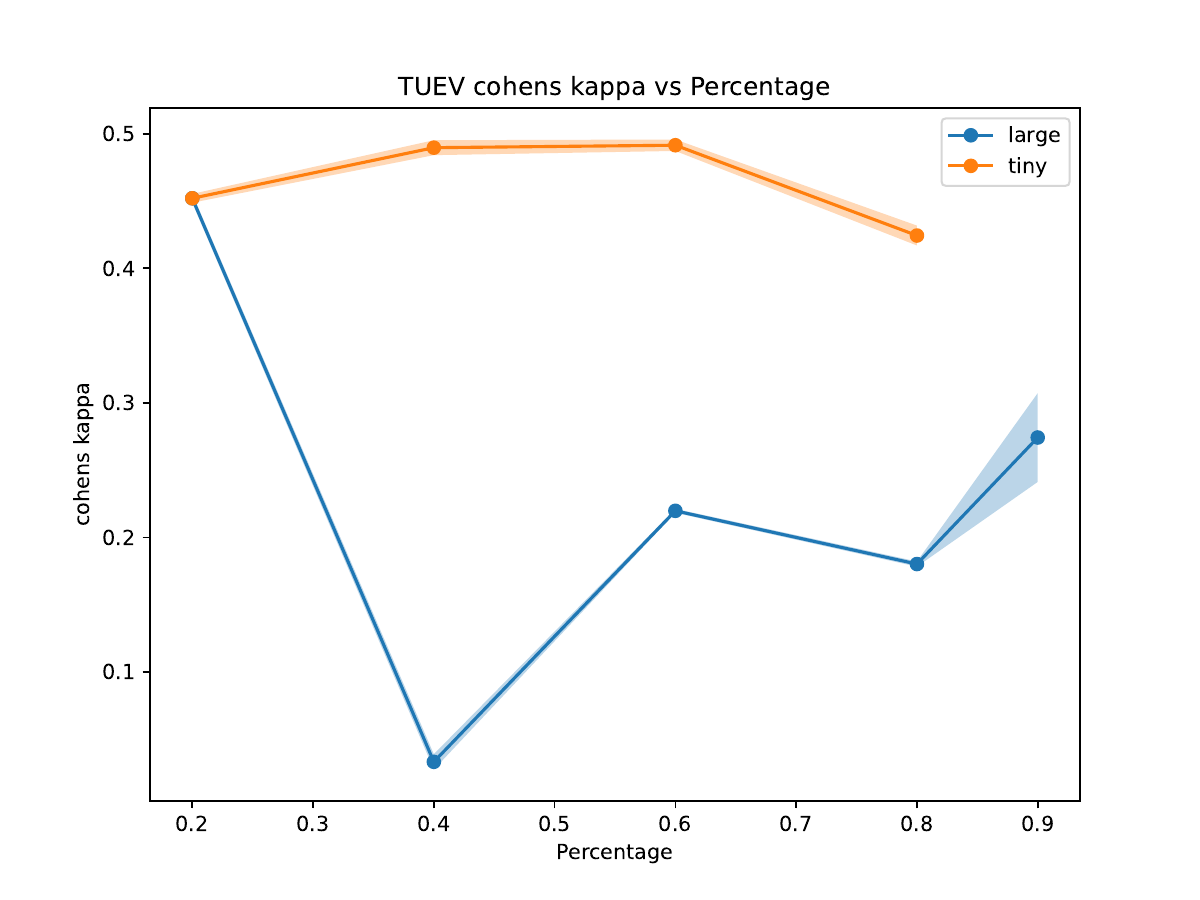}
  \label{fig:5}
\end{subfigure}\hfil 
\begin{subfigure}{0.33\textwidth}
  \includegraphics[width=\linewidth]{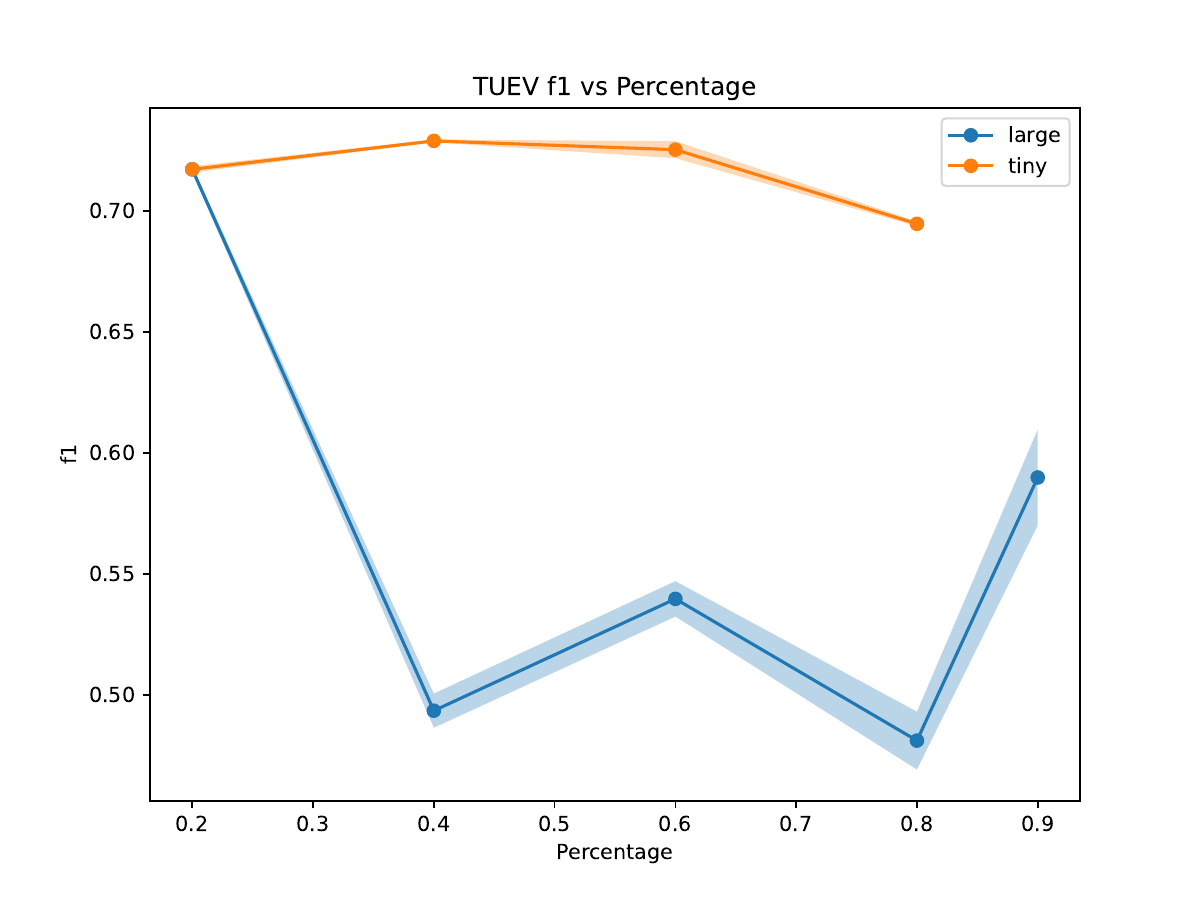}
  \label{fig:6}
\end{subfigure}}
\caption{A performance comparison on TUEV of MENDR Tiny and Large as pretraining data increases.}
\label{fig:scaling_images2}
\end{figure}

In \cite{LABRAM}, the authors questioned how much data is sufficient for training a large EEG model. We posit the complementary question: Can extremely small EEG foundation models scale with extensive data? Also, does scaling the number parameters improve performance? Since we have over 4,000 hours of EEG data, significantly larger than LaBraM’s 2,500 pretraining dataset, we can test the extremes of model size versus dataset size scaling. For these experiments, we pretrain the model on the pretraining dataset sampled at 128Hz, with only wavelet series from the $\delta, \theta, \alpha, \beta$ and $\gamma$ bands to establish the lower bound on the scaling properties of MENDR (scaling properties for larger models and larger frequency ranges are already shown in existing work like \cite{LABRAM, EEGPT}). In Figure \ref{fig:scaling_images1} and \ref{fig:scaling_images2}, we compare the MENDR Tiny and Large models after pretraining on different percentages (20\%, 40\%, 60\%, 80\%, 90\%) of the dataset and then evaluating the models on TUAB and TUEV. Note that we only assess the models without the "high" frequency band. Based on Figures \ref{fig:scaling_images1} and \ref{fig:scaling_images2}, it seems that pretraining is crucial to improving TUAB performance but not TUEV performance. 

\vspace{-3mm}
\subsection{Significantly Reducing the Number of Channels: ISRUC Dataset}
Unlike other EEG foundation models that can take varying input sizes, our work forces every dataset into the 19 specified channels. In effect, the autoencoder learns to impute missing channels if the dataset has fewer than 19 channels, such that the contextualizer always has a “whole brain view” as the dimensionality of the matrix embeddings must remain at 19. To demonstrate the efficacy of this approach, we utilize the ISRUC-SLEEP dataset, a sleep staging dataset comprising 100 all-night polysomnography recordings from 100 adults \cite{khalighi2016isruc}. Like \cite{CBRAMOD}, we only use the $6$ EEG channels (F3-A2, C3-A2, O1-A2, F4-A1, C4-A1, O2-A1) at 128 Hz sampling rate. Since these are bipolar electrodes, we approximate the $6$ as (F3, C3, O1, F4, C4, O2) for the purpose of electrode positioning. We only train the MENDR Tiny and Large on the common $\delta, \theta, \alpha, \beta, \gamma$, and our results are shown in Table \ref{tbl:ISRUCResultTable} with corresponding results from \cite{CBRAMOD}.  
\begin{table}[ht]
    \centering
    \caption{Results on ISRUC Dataset}
    \label{tbl:ISRUCResultTable}
    \resizebox{\linewidth}{!}{%
    \begin{tabular}{ccccc}
        \hline
        \hline
        & & \multicolumn{3}{c}{ISRUC (Multi-Class Classification)}\\
        \cmidrule(r){3-5}
        Methods & Model Size & Balanced Accuracy & Cohen’s Kappa & Weighted-F1 \\
        \hline
        \textbf{MENDR-Tiny} & 1.815M &  $0.6767 \pm 0.0108$ & $0.6256 \pm 0.0079$ & $0.7023 \pm 0.0065$ \\
        \textbf{MENDR-Large} & 1.817M & $0.5455 \pm 0.0239$ & $0.4508 \pm 0.0254$ & $0.5486 \pm 0.0199$ \\
        \hline
        SPaRCNet  & 0.79M & $0.7487 \pm 0.0075$ & $0.7097 \pm 0.0132$ & $0.7624 \pm 0.0092$ \\
        ContraWR & 1.6M & $0.7402 \pm 0.0126$ & $0.7178 \pm 0.0156$ & $0.7610 \pm 0.0137$\\
        CNN-Transformer  & 3.2M & $0.7363 \pm 0.0087$ & $0.7129 \pm 0.0121$ & $0.7719 \pm 0.0105$ \\
        FFCL  & 2.4M & $0.7277 \pm 0.0182$ & $0.7016 \pm 0.0291$ & $0.7614 \pm 0.0197$ \\
        ST-Transformer & 3.5M & $0.7381 \pm 0.0205$ & $0.7013 \pm 0.0352$ & $0.7681 \pm 0.0175$ \\
        BIOT & 3.2M &  $0.7527 \pm 0.0121$ & $0.7192 \pm 0.0231$ & $0.7790 \pm 0.0146$\\
        LaBraM-Base & 5.8M & $0.7633 \pm 0.0102$ & $0.7231 \pm 0.0182$ & $0.7810 \pm 0.0133$ \\
        CBraMod & 4.0M & $0.7865 \pm 0.0110$ & $0.7442 \pm 0.0152$ & $0.8011 \pm 0.0099$\\ 
        \hline
        \hline
    \end{tabular}%
    }
\end{table}
Note that, like \cite{CBRAMOD}, we break the sequence of 89420 30-second samples into 20 30-second chunks such that there are $4471$ datapoints in total. Unlike \cite{CBRAMOD}, we do not further split the training data into training and validation sets; instead, we use subjects 1 to 90 as the training set, while maintaining the same test set as \cite{CBRAMOD}. The models were fine-tuned over $8$ epochs with a learning rate of $0.0005$ and unfroze both the autoencoder and contextualizer, since neither module was pre-trained on data with this few channels. Unfortunately, it appears that both the tiny and large MENDR models underperformed several baselines; however, this could be due to insufficient training epochs/poor choice of hyperparameters. An interesting result is that the MENDR-Large performed worse than the MENDR-Tiny model despite having lower-dimensional embeddings/more layers. The comparison between the two models suggests that a lower-dimensional manifold is not sufficient for sleep stage prediction. 
\subsection{Manifold Explainability via UMAP}
\begin{figure}
    \centering
    \includegraphics[width=\linewidth]{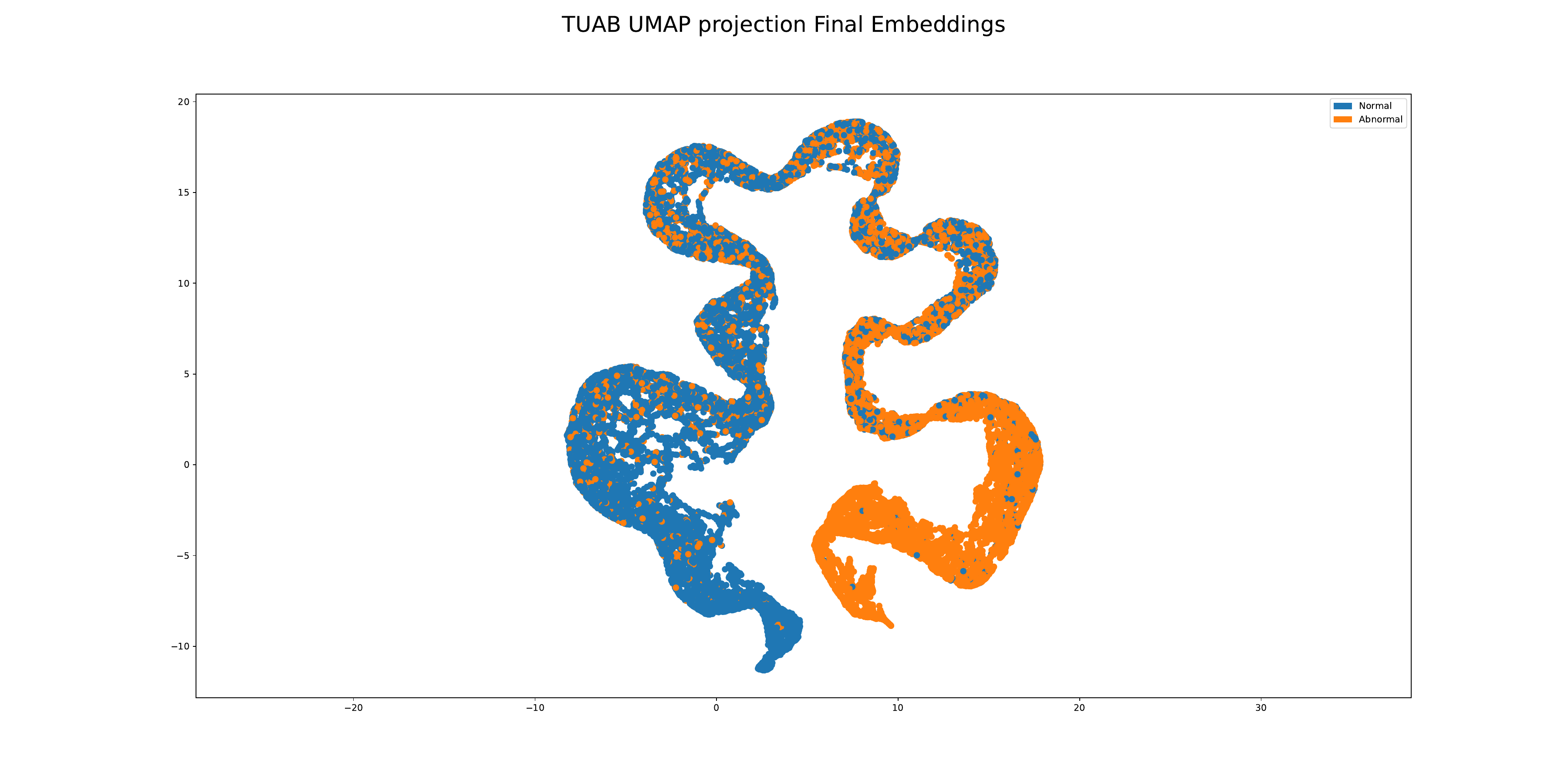}
    \caption{Learned UMAP Embeddings of TUAB}
    \label{fig:TUABUMAP}
\end{figure}

Uniform Manifold Approximation and Projection (UMAP) is a popular dimensionality reduction technique for visualizing data manifolds in lower dimensions \cite{mcinnes2018umap}. UMAP assumes that the data is uniformly distributed on a Riemannian manifold, the Riemannian metric is locally constant, and the manifold is locally connected. Although traditional work with Euclidean embeddings cannot entirely rely on UMAP for separating different classes due to these assumptions, MENDR’s construction and optimization on the Riemannian SPD manifold ensure that these assumptions hold. Hence, one can trust UMAP for explaining/visualizing the EEG SPD manifolds, as previously demonstrated in \cite{marin2025riemannian, zhang2023spatio}. For example, Figure \ref{fig:TUABUMAP} displays the UMAP visualization of the learned embeddings of the TUAB dataset, separated by class, immediately before being passed through the final linear transformation layer after the embeddings are projected back into Euclidean space from Riemannian space. Note that orange represents abnormal EEG patches, whereas blue represents normal EEG patches. More visualizations of the embeddings can be found in the appendix.

\section{Conclusion}
\label{sec:Conclusion}

In this work, we present the first EEG foundation model based on Riemannian SPD optimization. Although MENDR slightly underperforms or achieves baseline performance with various benchmark datasets, its performance-to-parameter ratio is higher than that of other state-of-the-art models. Future work will involve refining the manifold transformers to scale them up to larger dimensions and integrating multi-headed attention into the MAtt module.
\clearpage
\bibliography{ref}
\clearpage
\appendix
\label{sec:Appendix}
\section{Appendix}
\subsection{Related Works}

\paragraph{EEG Foundation Models.} The current state-of-the-art method for EEG foundation model architecture design is Edisonian: existing studies each perform trial-and-error search on larger and larger models in hopes of developing better EEG representations. For example, after BENDR \cite{BENDR}, which included 0.39M parameters, the authors of \cite{zhang2023brant} proposed Brain Neural Transformer (Brant), an EEG foundation model with 500M parameters, and Brant-2 \cite{brant2}, with over 1 billion parameters. LaBraM proposed the LaBraM-Huge \cite{LABRAM} with 369M parameters and NeuroLM \cite{https://doi.org/10.48550/arxiv.2409.00101} with 1.7B parameters, the authors of \cite{FoME} proposed the Foundation Model for EEG (FoME) with 745M parameters, the authors of \cite{wang2023brainbert} proposed BrainBERT with 43.18M parameters, the authors of \cite{CBRAMOD} proposing CBraMod with 4M parameters (although their decoders for several of their downstream tasks are of the size of 100M parameters), and the authors of \cite{EEGPT} propose EEG Pre-trained Transformer (EEGPT) with over 10M parameters. All models mentioned use embedding reconstruction losses, sometimes the masked autoencoding paradigm proposed by \cite{VIT}, or an autoregressive reconstruction framework proposed in EEGPT. Note that \cite{LABRAM} splits the pre-training phase into two phases: generating embeddings using a neural tokenizer and decoder first to generate vector quantized neural spectrum embeddings and the second phase feeds these embeddings into a transformer block. MENDR instead combines both phases into one phase during pre-training to maintain \textit{temporal explainability of the embeddings and manages to reconstruct temporal domain information in the form of wavelets, in contrast to only frequency domain information in LaBraM's tokenizer}.

\paragraph{Wavelet Theory in Self-Supervised EEG.} \cite{wavelet2vec} proposed the Wavelet2Vec framework for seizure subtype classification that proposes a filter-bank architecture similar to MENDR; however, after decomposing their input of $X \in \mathbb{R}^{C \times 1 \times N}$, where $C$ is the number of channels and $N$ is the number of time-steps, into $X_{wavelet} \in \mathbb{R}^{6 * \times C \times 1 \times N}$, where $6$ represents the $5$ frequency bands used in MENDR and the “high” frequency band (64-128 Hz), they treat the input as an image and feed it into a Vision Transformer (ViT) \cite{VIT2}. We excluded the high band because of our band-pass filtering between 0.1 and 75 Hz. BrainBERT \cite{wang2023brainbert} proposed using the superlet transform \cite{Moca2021}, a composition of Morlet wavelet transforms that is more Pareto optimal than just a Morlet wavelet transform concerning the tradeoff between time and frequency resolution. The use of the superlet transform by BrainBERT was in response to how other EEG foundation model studies, \cite{BIOT, LABRAM, CBRAMOD}, used short-time Fourier Transforms (STFTs) to extract frequency domain features from their EEG signals with temporal localization, and does not explicitly correspond to human brain rhythms. Something to note is that the Discrete Wavelet Transform has $O(N)$ complexity, whereas calculating the STFT is $O(N\log N)$, and our filter bank decomposition method calculates $5$ levels of decomposition, which means $5$ wavelet transform calculations, whereas the superlet transform proposed by \cite{wang2023brainbert} is the geometric mean of $3-30$ Morlet wavelet transforms (depending on the order parameter, which fluctuates based on their proposed adaptive order method). To our knowledge, the only other EEG foundation model study that uses wavelet theory is Beatrix \cite{zheng2025beatrix}. The authors in Beatrix employ the Analytic Wavelet Transform (AWT) \cite{lilly2010analytic} similar to the superlet transform in \cite{wang2023brainbert}, which is designed to provide a more accurate estimation of the instantaneous frequency and superior frequency reassignment properties compared to the DWT and STFT, and also try to develop a more parameter-efficient EEG foundation model, though specifics on the model size with respect to the AWT is unknown due to lack of published code. 

\paragraph{GNN-based EEG Networks} The use of GNNs for deep learning of EEG signals is an emerging field \cite{klepl2024graph}. 

\paragraph{Riemannian Geometry-Based EEG Learning.} The application of Riemannian Geometry to EEG-based Brain Computer Interface (BCI) decoding is an emerging field that falls under the broader category of manifold/constrained optimization \cite{tibermacine2024riemannian}. The most common Riemannian manifold studied in EEG is the SPD manifold, which inspired the Manifold Attention (Matt) module from \cite{pan2022matt} that we use in our Manifold Transformer for MENDR. Due to SPD matrix embeddings of patches lying not in a vector space but a Riemannian manifold, existing metrics for learning based on Euclidean structure cannot be applied to SPD matrix embeddings \cite{huang2015log}. The Affine-Invariant Metric (AIM) proposed by \cite{pennec2006riemannian} was one of the first metrics for measuring the similarity between matrix embeddings on the SPD manifold, but was superseded by the Log-Euclidean Metric (LEM) \cite{arsigny2006log, arsigny2007geometric} due to its computational efficiency and bi-invariance on the Lie group of SPD matrices. In classical unconstrained optimization, the search space is linear (in Euclidean space $\mathbb{R}^{n}$, and we want to optimize the cost function:

\begin{equation}
    \min_{x \in \mathbb{R}^{n}} f(x)
\end{equation}

and use the dot product as a measure of similarity in Euclidean space: $\langle u, v \rangle = u^{\top}v$. When $f(x)$ is differentiable, one can calculate the gradients $\nabla f(x)$ and Hessians $\nabla^{2} f$ of $f(x)$, which underpins why gradient descent is more efficient than evolutionary random search algorithms for optimization. Manifold optimization takes gradient descent a further step under the assumption that the search space is instead a smooth manifold, thereby constraining the optimization \cite{boumal2023introduction, absil2009optimization}:

\begin{equation}
    \min_{x \in \mathcal{M}} f(x)
\end{equation}

and use the LEM as a measure of similarity along $\mathcal{M}$ and using Riemannian gradients and Hessians for optimization. Motivated by the efficiency of manifold optimization due to evidence suggesting neural activity lies on lower-dimensional manifolds \cite{Langdon_2023, can2021emergence}, we designed MENDR to “kill two birds with one stone”: a computationally efficient EEG foundation model that aligns with existing neuroscience studies. Riemannian geometry-based learning of EEG is still a very nascent field, \cite{zhang2023spatio, tibermacine2024riemannian}, and thus very little work has been done on Riemannian self-supervised learning and learning generalized EEG embeddings on the Riemannian manifold. Specifically, given the powerful ability of attention to generate generalized representations of input data as demonstrated in \cite{vaswani2017attention}, to our knowledge, the only other works that apply manifold attention to EEG besides \cite{pan2022matt} are \cite{lu2024manifold, qin2024bnmtrans}, and these studies specifically focus on specific downstream tasks rather than generating generalized EEG representations. 

\subsubsection{Data Preprocessing}
\label{subsec:DataPreprocess}
Data preprocessing is crucial to MENDR and is the primary reason we can achieve a good performance-to-parameter ratio.
\subsubsection{Raw EEG Signal Preprocessing.} To begin, similar to \cite{CBRAMOD}, we remove EEG recordings with a duration of no longer than three minutes in our pretraining dataset, and the first minute and last minute of each recording are removed, as we noticed significant noise/artifacts in the beginning and end of the EEG recordings. In downstream tasks, we do not discard recordings due to short length and do not remove the first and last minutes. Next, we select the nineteen electrodes in the international 10-20 system, excluding the anterior lobe electrodes: (Fp1, Fp2, F7, F3, Fz, F4, F8, T3, C3, Cz, C4, T4, T5, P3, Pz, P4, T6, O1, O2). We argue that nineteen is a special number for EEG harmonization because the nineteen electrode positions in the 10-20 system are designed to measure the entirety of the brain’s electrical activity as best as possible on the scalp. Next, we apply a band-pass filter between 0.1 Hz and 75 Hz to remove low-frequency and high-frequency noise, and a notch filter at 60 Hz and its harmonics if the sampling rate is greater than 240 Hz. Afterwards, we resample the recordings to \textbf{128 Hz and segment them into 1-minute non-overlapping EEG samples}. We choose $128$ Hz rather than the traditional $200$ Hz \cite{BIOT, LABRAM, CBRAMOD} because $128 = 2^{6}$, which nicely defines the frequency band splits in the wavelet decompositions. Then, since EEG is usually recorded in $\mu$V, we multiply the EEG values by $10^{5}$ to prevent future underflow errors. Finally, in the pretraining dataset, we limit the total length of recordings for a patient to one hour to avoid bias from overfitting the data to a single patient. The modified TUEG pretraining dataset contains 259,712 EEG samples and represents over 4,328 hours of EEG, larger than the amount of EEG samples used for pre-training LaBraM ($\sim$ 2,535 hours) \cite{LABRAM}.

\subsubsection{Discrete Wavelet Packet Transform (DWPT).} In contrast to raw signal analysis or frequency-domain Fourier analysis of EEG signals, wavelet analysis balances the two domain resolutions to simultaneously localize information in both domains. In EEG analysis, the most widely studied bands related to human brain rhythms are the $\delta$ ($0$ - $4$ Hz), $\theta$ ($4$ - $8$ Hz), $\alpha$ ($8$-$16$ Hz), $\beta$ ($16$-$32$ Hz), and $\gamma$ ($32$ - $64$) frequency bands \cite{ABHANG201619}. We decompose every EEG signal into these $5$ frequency bands by calculating the Discrete Wavelet Transform (DWT) and extracting the relevant approximation and detail coefficients from the DWPT tree in Appendix Figure \ref{fig:WaveletPacketDecomposition}.
\subsubsection{EEG Graph Creation.} In MENDR, we use a Graph Attention Convolutional Network (GATConv) \cite{GAT} to learn a deep spatial harmonization of the EEG data. In a GNN, we specify a fixed weight between certain tokens/nodes in a grouping of embeddings through the edge weights. Ultimately, this introduces a human bias into the “attention” of GNNs, and one can interpret this bias as a non-random initialization of the attention values between different EEG channels for spatial encoding. Specifically, based on the 10-20 system specified in the dataset, we can derive 3D coordinates for the electrodes. Using the most popular MNI (Montreal Neurological Institute) coordinate system as our reference, we model each subject’s head as a sphere with radius normalized to $1$ \cite{wu2018accurate}. Note that a sphere is one of the most common examples of a Riemannian manifold, and the geodesic is the shortest distance between two points on that manifold. Thus, given the 3D MNI coordinates of two channels $i$, ($x_{i}$, $y_{i}$, $z_{i}$) and $j$, $(x_{j}, y_{j}, z_{j})$, the geodesic distance $D_{ij}$ along a sphere’s radius $r$ is:

\begin{align}
    D_{ij} = arccos(\frac{x_{i}x_{j} + y_{i}y_{j} + z_{i}z_{j}}{r^{2}})
    \label{eq:Dist}
\end{align}

In cases where electrode coordinates are specified and one cannot assume the head is a sphere with a radius of 1, we normalize the distances to a range of 0 to 1. As spline interpolation has taught us \cite{Moguilner2024}, electrode positions provide sufficient information for reconstructing missing channels in contrast to the multitude of potential edge features used in GNN-based EEG models \cite{klepl2024graph, wagh2020eeg}. In summary, we create five graphs for each EEG signal, each corresponding to the graph representation of the wavelet coefficients.

\textbf{Patchifying Wavelet Data.} Like previous EEG and time series foundation model studies, such as CBraMod \cite{CBRAMOD}, LaBraM \cite{LABRAM}, and the patch time series transformer (PatchTST) \cite{PatchTST}, the Masked Autoencoder paradigm \cite{VIT} has proven empirically to be an effective self-supervised pretraining task. However, due to the nature of wavelet decompositions, frequency band decompositions of the EEG have varying time samples. Additionally, unlike the patch-based learning presented in these works, our embeddings ultimately become SPD matrices, whose size is determined by the number of channels. As mentioned before, we cannot use one-second-long patches because wavelet decomposition reduces the temporal resolution of EEG signals. Instead, we use \textbf{two-second non-overlapping patches}. Each band decomposition is divided into the same number of patches, where the number of sample points depends on the frequency band; readers are referred to \ref{table:wavelet-patch-lengths} for more details. For example, in the case of 1-minute segments in the pre-training dataset, we deconstruct the segment into 30 2-second segments. 

\subsection{Mathematical Preliminaries}
\label{subsec:Mathematical Preliminaries}

\subsubsection{Symmetric Positive Definite Matrix Learning}

A matrix $\mathbf{A} \in \mathbb{R}^{n \times n}$ is Symmetric Positive Definite (SPD) if $\mathbf{A} = \mathbf{A}^{T}$ and $x^{T}Ax > 0$ for all non-zero vectors $x \in \mathbb{R}^{n}$. The eigenvalues of $\mathbf{A}$, $\lambda_{1}, \ldots, \lambda_{n}$ are all guaranteed to be positive. The exponential and logarithmic operators on $\mathbf{A}$, denoted as exp($\mathbf{A}$) and log($\mathbf{A}$), are defined through their orthogonal diagonalization. If $\mathbf{A}$ is expressible as $\mathbf{U}$diag($\lambda_{1}, \ldots, \lambda_{n})\mathbf{U}^{T}$, where $U$ are the orthogonal eigenvectors of $\mathbf{A}$, then 

\begin{equation}
    \text{exp}(\mathbf{A}) = \mathbf{U}diag(exp(\lambda_{1}), \ldots, exp(\lambda_{n}))\mathbf{U}^{T}
    \label{eq:ExpMatrix}
\end{equation}

and

\begin{equation}
    \text{log}(\mathbf{A}) = \mathbf{U}diag(log(\lambda_{1}), \ldots, log(\lambda_{n}))\mathbf{U}^{T}
    \label{eq:LogMatrix}
\end{equation}

The SPD manifold is Riemannian, but calculating the Riemannian mean of a set of SPD matrices has no closed-form solution. Let $\mathcal{M}$ denote a Riemannian manifold and $A, B \in \mathcal{M}$. In Manifold Attention, the Riemannian mean is approximated using the Log-Euclidean metric (LEM) $\delta_{L}(A, B)$ as an approximation of the geodesic distance between $A$ and $B$: 

\begin{align}
    \delta_{L}(A, B) = \|\text{log}(A) - \text{log}(B)\|_{F}
\end{align}

And the Log-Euclidean mean, which does have a closed form solution and is SPD (see \cite{pan2022matt}), $\mathcal{G}$ as:

\begin{align}
    \mathcal{G} = \text{exp}(\frac{1}{k} \sum_{l=1}^{k} \text{log}(A_{l}))
\end{align}

For an EEG signal/wavelet decomposition, consider $\mathbf{X} \in \mathbf{R}^{C \times T}$ and let $SCM$ be the sampled covariance matrix:

\begin{align}
    SCM = \frac{1}{T - 1}XX^{T}
\end{align}

To prove why this is SPD, we ask readers to consult \cite{tibermacine2024riemannian}. 

\subsubsection{MAtt Attention Module}
We utilize the Manifold Attention (MAtt) module proposed by \cite{pan2022matt} to perform attention calculations on the SPD manifold. To first understand the MAtt module, we have first to understand the BiMap layer proposed by \cite{huang2017riemannian}:

\begin{equation}
    \mathbf{A_{k}} = f_{\text{BiMap}}^{(k)}(\mathbf{A}_{k - 1}; \mathbf{W}_{k}) = \mathbf{W}_{k} \mathbf{A}_{k-1} \mathbf{W}_{k}^{T}
\end{equation}

where $\mathbf{A_{k}}$ is an SPD matrix at layer $k$ of the network, $\mathbf{W}_{k} \in \mathbb{R}_{*}^{d_{k} \times d_{k - 1}}$, the $*$ means $\mathbf{W}_{k}$ must be full rank and $d_{k} \leq d_{k - 1}$, but we only use the $d_{k} = d_{k - 1}$ case. Optimizing $\mathbf{W}_{k}$ is non-trivial and requires optimizing it within a Stiefel manifold. We recommend that readers interested in learning more refer to the original work proposed by SPDNet \cite{huang2017riemannian}. We use the BiMap layer analogously to the Feed-Forward Network in the original transformer architecture \cite{vaswani2017attention}. Then, we modularize equation \eqref{eq:BatchTraceNorm} into a BatchTraceNorm module that trace-normalizes each sample in the batch, using equation \eqref{eq:BatchTraceNorm} as an analogue to LayerNorm in the original transformer encoder implementation \cite{vaswani2017attention}. Then attention is calculated using the BiMap Layer: 

in manifold attention, the weights of for calculating the query, key, and value $W_{q}, W_{k}, W_{v}$ of input $x_{i}$ must lie on a Stiefel manifold for query $q_{i}$, key $k_{i}$, and value $v_{i}$ to be SPD:

\begin{align}
    q_{i} &= W_{q}x_{i} W_{q}^{T} \\
    k_{i} &= W_{k}x_{i} W_{k}^{T} \\
    v_{i} &= W_{v}x_{i} W_{v}^{T} 
\end{align}

Without loss of generality, consider only $W_{v}$ and the calculated Euclidean gradient from traditional backpropagation as $\nabla_{W_{v}} \mathcal{L}$. To calculate the Stiefel gradient, $\Delta_{W_{v}} \mathcal{L}$ is projected onto the normal/orthogonal space of the tangent space of $W_{v}$, with the projection denoted as $\pi_{N}(\nabla_{W_{v}} \mathcal{L}) = W_{v}(W_{v}^{T}\nabla_{W_{v}} \mathcal{L})_{sym}$

where $X_{sym} = \frac{X + X^{\top}}{2}$. Then the tangent component of the Stiefel gradient for $W_{v}$ can be defined as the subtraction between the Euclidean gradient and $\pi_{N}(\nabla_{W_{v}} \mathcal{L})$

\begin{equation}
   \hat{\nabla}_{W_{v}} \mathcal{L} = \Delta_{W_{v}} \mathcal{L} - \pi_{N}(\nabla_{W_{v}} \mathcal{L})
\end{equation}

After moving along the Stiefel Gradient, we cannot guarantee that the weights remain on the Stiefel manifold. So we have projected them onto the Stiefel manifold through a retraction operation:

\begin{equation}
    W_{v}^{(new)} = \Gamma(W_{v} - \eta  \hat{\nabla}_{W_{v}} \mathcal{L})
\end{equation}

where $\Gamma$ is the retraction operation defined in QR decomposition. We ask readers \cite{pan2022matt, huang2017riemannian, absil2009optimization} for a more rigorous derivation. Note that, due to numerical instability, we also perform regularization, i.e., $X_{sym}$, before outputting the SPD patch embeddings in the manifold transformer.

\subsubsection{Riemannian Residual Networks}

Because we are manipulating embeddings along a Riemannian manifold, we cannot use traditional residual connections in our architecture. Instead, by \cite{katsman2023riemannian}, we must use Riemannian residuals. Since we define the manifold with the Log-Euclidean metric, given matrix embedding $\mathbf{A}$ and $\mathbf{B}$, the residual connection output $\hat{\mathbf{A}}$ is defined as:

\begin{equation}
    \hat{\mathbf{A}} = \exp(\log(\mathbf{A}) + \log(\mathbf{B}))
\end{equation}

where $\log$ and $\exp$ are the matrix logarithm and exponential, respectively. The residual connection can be interpreted as projecting the matrix embeddings back into Euclidean space, performing the addition of the embeddings in Euclidean space, and then re-projecting the result back into Riemannian space.

\section{Numerically Stable SVD Differentiation}

\cite{townsend2016differentiating} proposed differentiating the Singular Value Decomposition concerning an optimization loss, which is implemented in PyTorch’s Autograd. Although SVD is defined for any $m \times n$ matrix, we can skip certain gradient computations to accelerate backpropagation because we only deal with square $n \times n$ matrices. However, we cannot guarantee that the matrix to be eigendecomposed is symmetric due to inaccuracies in floating-point representation. Moreover, EEG data has a low signal-to-noise ratio, making it difficult to differentiate the constructed SPD matrices. 

Specifically, for a full rank square matrix $\mathbf{A} \in \mathbb{R}^{n \times n}$ the SVD eigendecomposition is defined as: $\mathbf{A} = \mathbf{U}\mathbf{S}\mathbf{V}^{\top}$, where $\mathbf{U} \in \mathbf{R}^{n \times n}$, $S \in \mathbf{R}^{n \times n}$ is a diagonal matrix containing the singular values of $\mathbf{A}$, and $V \in \mathbb{R}^{n \times n}$. Additionaly, $\mathbf{U}$ and $\mathbf{V}$ are orthonormal. When $\mathbf{A}$ is symmetric, $\mathbf{U} = \mathbf{V}$. As described in Townsend \cite{townsend2016differentiating}, the differential of $\mathbf{A}$ is: 

\begin{equation}
    d \mathbf{A} = d \mathbf{USV}^{\top} + \mathbf{U}d\mathbf{SV}^{\top} + \mathbf{US}d\mathbf{V}^{\top}
    \label{eq:SVDDiff}
\end{equation}

Given the constraint that $\mathbf{U, V}$ are orthonormal, the $d\mathbf{U}$ and $d\mathbf{V}$ are constrained by the following equations:

\begin{equation}
    d\mathbf{U}^{\top}\mathbf{U} + \mathbf{U}^{\top}d\mathbf{U} = \mathbf{0}
\end{equation}

\begin{equation}
    d\mathbf{V}^{\top}\mathbf{V} + \mathbf{V}^{\top}d\mathbf{V} = \mathbf{0}
\end{equation}

Let $d\Omega_{\mathbf{U}} =  d\mathbf{U}^{\top}\mathbf{U}$ and $d\Omega_{\mathbf{V}} = d\mathbf{V}^{\top}\mathbf{V}$. Note that $d\Omega_{\mathbf{U}}$ and $d\Omega_{\mathbf{V}}$ are \textit{skew-symmetric}, meaning that it is equal to the negative transpose of themselves. From Edelman \cite{edelman1998geometry}, if we fix orthogonal matrices $\mathbf{U}_{\bot}, \mathbf{V}_{\bot}$ such that $[\mathbf{U} \mathbf{U}_{\bot}]$ and $[\mathbf{V} \mathbf{V}_{\bot}]$ (which can be found from the Grad-Schmidt process), we may then expand $d\Omega_{\mathbf{U}}$ and $d\Omega_{\mathbf{V}}$ as:

\begin{equation}
    \mathbf{dU} = \mathbf{U}d\Omega_{\mathbf{U}} + \mathbf{U}_{\bot}d\mathbf{K_{\mathbf{U}}}
\end{equation}

\begin{equation}
    \mathbf{dV} = \mathbf{V}d\Omega_{\mathbf{V}} + \mathbf{V}_{\bot}d\mathbf{K_{\mathbf{V}}}
\end{equation}

where $d\mathbf{K_{U}}$ and $d\mathbf{K_{V}}$ are unconstrained matrices. 

Now, by left multiplying equation \eqref{eq:SVDDiff} $\mathbf{U}^{\top}$ and right multiplying by $\mathbf{V}$ we get:

\begin{equation}
    \mathbf{U}^{\top}\mathbf{AV} = d \Omega_{\mathbf{U}}\mathbf{S} + d\mathbf{S} + \mathbf{S}d\Omega^{\top}_{\mathbf{V}}
\end{equation}

If we let $d\mathbf{P}:= \mathbf{U}^{\top}\mathbf{AV}$, $\bar{\mathbf{I}}_{k}$ denote the $n \times n$ matrix with zero diagonal and ones everywhere else, and having $\circ$ denote the Hadamard product, the diagonal and off diagonal can be computed as:

\begin{equation}
    d\mathbf{S} = \mathbf{I}_{k} \circ d \mathbf{P}
\end{equation}

\begin{equation}
    \bar{\mathbf{I}}_{k} \circ d \mathbf{P} = d \Omega_{\mathbf{U}} \mathbf{S} - \mathbf{S}d\Omega_{\mathbf{V}}
    \label{eq:OffDiag}
\end{equation}

For more details on how these equations were derived, we refer readers to \cite{townsend2016differentiating}. The transpose of the left-hand side of \eqref{eq:OffDiag} yields: 

\begin{equation}
     \bar{\mathbf{I}}_{k} \circ d \mathbf{P}^{\top} = -\mathbf{S} d\Omega_{\mathbf{U}} + d\Omega_{\mathbf{V}}\mathbf{S}
     \label{eq:OffDiagTranspose}
\end{equation}

Afterwards, right multiply equation \eqref{eq:OffDiag} by $\mathbf{S}$ and left multiply \eqref{eq:OffDiag} by $\mathbf{S}$ and add, yielding:

\begin{equation}
    \bar{\mathbf{I}}_{k} \circ \Bigg[ d\mathbf{PS} + \mathbf{S}d\mathbf{P}^{\top} \Bigg] = d \Omega_{\mathbf{U}}\mathbf{S}^{2} - \mathbf{S^{2}}d\Omega{\mathbf{U}}
\end{equation}

which is solved by:

\begin{equation}
    d\Omega_{\mathbf{U}} = \mathbf{F} \circ \Bigg[ d\mathbf{PS} + \mathbf{S}d\mathbf{P}^{\top} \Bigg]
\end{equation}

For $d\Omega_{\mathbf{V}}$, by an identical process we have:

\begin{equation}
    d\Omega_{\mathbf{V}} = \mathbf{F} \circ \Bigg[ d\mathbf{PS} + \mathbf{S}d\mathbf{P}^{\top} \Bigg]
\end{equation}

Here $\mathbf{F}$ represents the matrix:

\begin{equation}
    \mathbf{F}_{ij} = \begin{cases} 
      \frac{1}{s_{j}^{2} - s_{i}^{2}} &  i \neq j \\
      0 & i = j \\
   \end{cases}
\end{equation}

Herein lies the problem: if the singular values of $\mathbf{A}$ are close to each other, the values of the diagonal elements in $\mathbf{F}$ explode, causing the exploding gradients problem. Due to the risk of EEG data being obstructed, noisy, and stochastic, this problem occurred frequently during our training. We resolve this problem identical to \cite{PhysRevX.9.031041} by adding a small $\epsilon = 1 \times 10^{-12}$ to the denominator term:

\begin{equation}
    \mathbf{F}_{ij} = \begin{cases} 
      \frac{\frac{1}{s_{j}^{2} - s_{i}^{2}}}{(s_{j}^{2} - s_{i}^{2})^{2} + \epsilon} &  i \neq j \\
      0 & i = j \\
   \end{cases}
\end{equation}

In the symmetric case, i.e., using \texttt{torch.eigh} instead of \texttt{torch.svd}, we simply add a smaller $\epsilon = 1 * 10^{-20}$ like \cite{PhysRevX.9.031041}:

\begin{equation}
    \mathbf{F}_{ij} = \begin{cases} 
      \frac{1}{({s_{j}^{2} - s_{i}^{2}}) + \epsilon} &  i \neq j \\
      0 & i = j \\
   \end{cases}
\end{equation}

To derive $d\mathbf{K_{U}}$ we left multiply \eqref{eq:SVDDiff} by $\mathbf{U}_{\bot}^{\top}$ which gives and then implies equation \eqref{eq:dKU}:

\begin{equation}
    \mathbf{U}^{\top}_{\bot}d\mathbf{A} = d \mathbf{K_{U}SV^{\top}}
\end{equation}

\begin{equation}
    d\mathbf{K_{U}} = \mathbf{U}_{\bot}^{\top} d \mathbf{AVS^{-1}}
    \label{eq:dKU}
\end{equation}

By a similar line of reasoning for $\mathbf{V}$:

\begin{equation}
    d\mathbf{K_{V}} = \mathbf{V}_{\bot}^{\top} d \mathbf{A^{\top}US^{-1}}
    \label{eq:dKV}
\end{equation}

These can now all be combined into formulas for the differentials $d\mathbf{U}$, $d\mathbf{S}$, $d\mathbf{V}$ in terms of $d\mathbf{A}$, $\mathbf{U}$, $\mathbf{S}$, and $\mathbf{V}$. For the explicit solutions, refer to \cite{townsend2016differentiating}. Potential future work would involve experimenting with different fast and approximate methods for differentiating the SVD, such as approximating the SVD differentials using a Taylor expansion \cite{wang2021robust}.

\section{Discrete Wavelet Packet Decomposition (DWPT)}

Suppose we have a discrete-time signal $x[t] = [x_{0}, x_{1}, \ldots, x_{t}]$. The Discrete Wavelet Transform (DWT) is a dyadic discretization of the scale parameter $a$ such that the interval $[a, b] = [2^{j}, k2^{j}]$, where $j$ represents the Level of decomposition and $k$ represents number of points to skip (i.e. subsampling). Then, a discrete-time function $x(t)$ can be reconstructed from mathematical manipulation of these two functions (see \cite{WaveletOG} for the mathematical manipulation):

\begin{align}
    \psi_{j, k}[t] = 2^{\frac{j}{2}} \sum_{k} d_{j,k} \psi [2^{j}t - k] \\
    \phi_{j, k}[t] = 2^{\frac{j}{2}} \sum_{k} a_{j,k} \phi [2^{j}t - k]
    \label{eq:DWT}
\end{align}
    
$d_{j, k}$ and $a_{j, k}$ are the high-frequency/detail and low-frequency/approximation coefficients, respectively. Also, the $\phi$ function is a scaling function derived from the chosen mother wavelet. Thus, one can think of the wavelet decomposition as breaking down a signal into high- and low-frequency components using high-pass and low-pass filters. In this work, we choose the Daubechies-4 (db4) family of wavelets because it is the most frequently used in existing EEG wavelet analysis literature, which specifies the definitions of $\psi$ and $\phi$ \cite{wavelet2vec}. The Discrete Wavelet Packet Transform (DWPT) is simply calculating both child nodes of the decomposition tree rather than just the approximation nodes, as shown in Appendix Figure \ref{fig:WaveletPacketDecomposition}.

\begin{figure*}
  \centering
  \includegraphics[width=\textwidth]{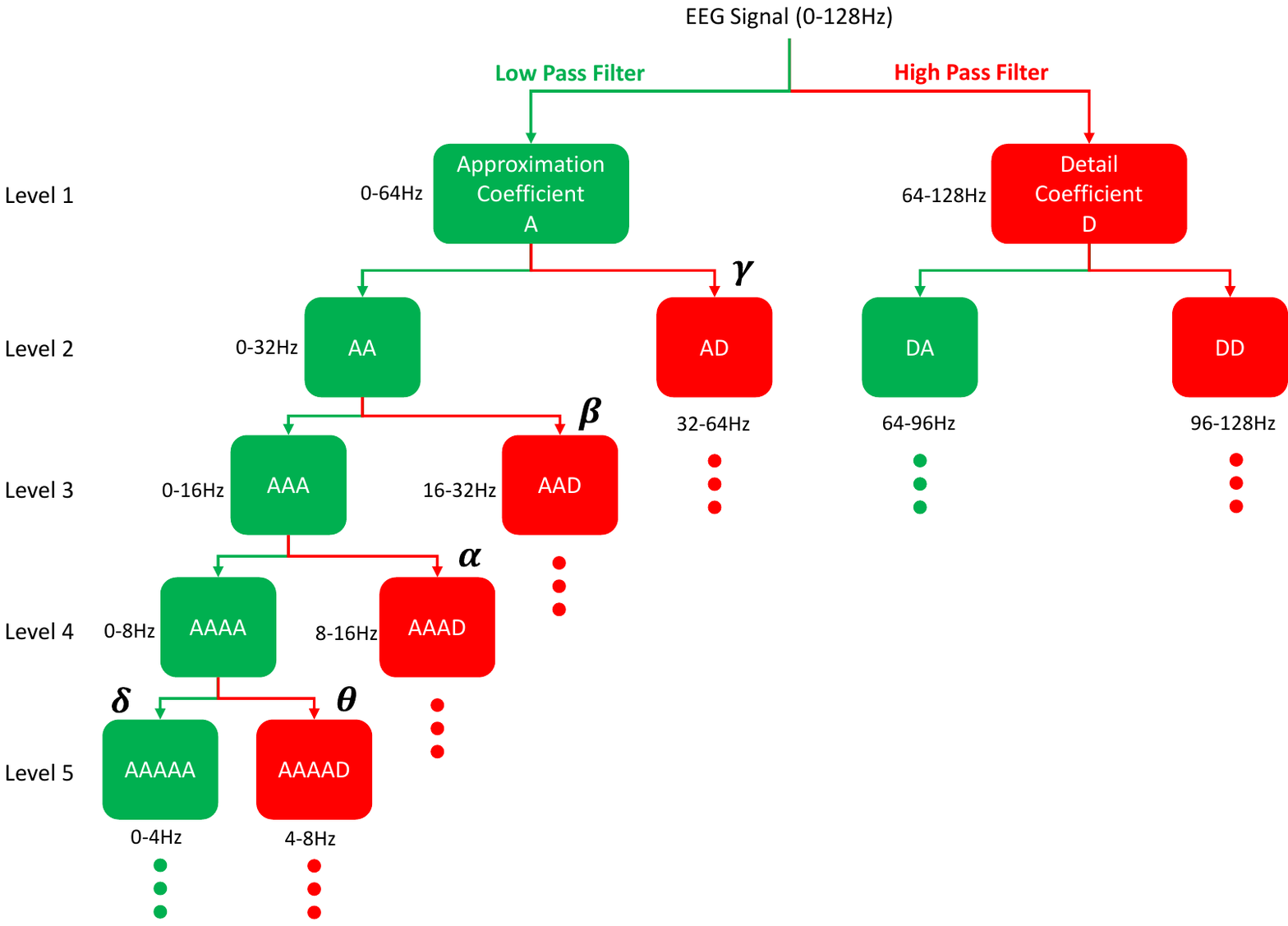}
  \caption{The Wavelet Packet Decomposition. Note that certain nodes in the tree represent specific human brain rhythms, which is why we used the wavelet transform. Inside each node, the labeling represents the path taken to reach that node through decomposition. For example, “AAAD” means the node is Level $4$ because there are $4$ characters and the parent node, on Level $3$, must be “AAA”. Additionally, the “D” after the “AAA” means that it is the detail coefficient of “AAA” and is typically denoted on the right-hand side of a binary decomposition tree. Similarly, the “DD” must have a parent of “D” and represents the detail coefficients of the wavelet decomposition of the detail coefficients of the original EEG signal. Also note that the children nodes have half the frequency range of the parent nodes, with the lower half associated with the approximation coefficients and the upper half associated with the detail coefficients.}
  \label{fig:WaveletPacketDecomposition}
\end{figure*}

\section{Module Architecture Diagrams} In Appendix Figures \ref{fig:MENDREncoder} and \ref{fig:MENDRDecoder}, we present the architectures for the autoencoder’s encoder and decoder. In Appendix Figure \ref{fig:MAtt Architecture} we present the architecture of the original Manifold Attention (MAtt) module architecture presented by \cite{pan2022matt}. In Appendix Figures \ref{fig:MENDRContextualizerTiny} and \ref{fig:MENDRContextualizerLarge}, we present more detailed architectures of both the tiny and large contextualizers. Note that our trace normalization is identical to the original trace normalization presented by \cite{pan2022matt} when constructing the original SPD matrices.

\begin{figure}
    \centering
	\includegraphics[width=0.5\textwidth]{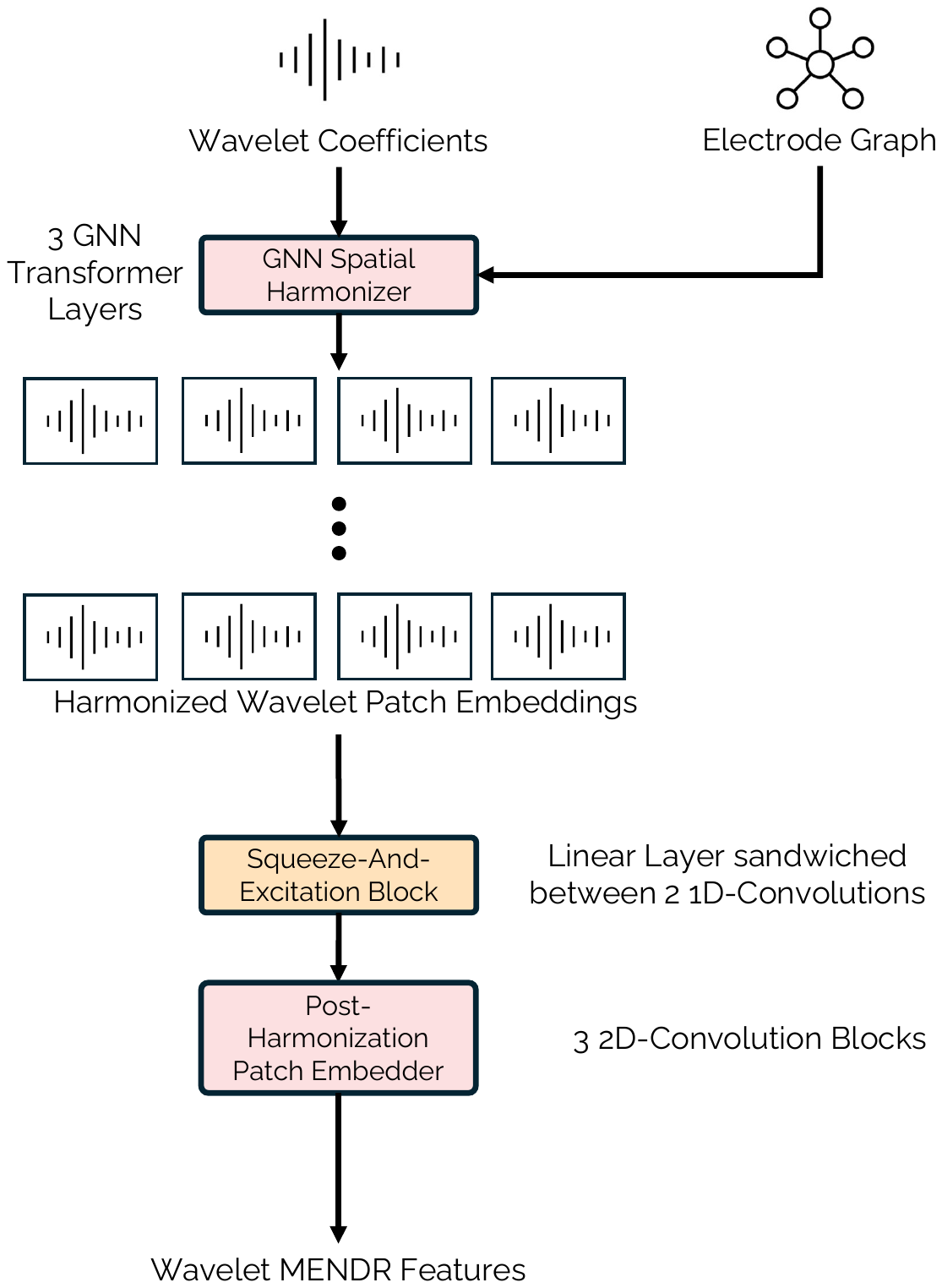}
	\caption[MENDR Encoder Architecture]{The MENDR Encoder module architecture. Note that the GNN module used is the Graph Attention Network module from \cite{GAT}. The Squeeze and Excitation block employs 1D convolutions along the channel dimension and does not reduce the channels as they pass through the convolutional layers (i.e., the reduction parameter is set to 1).  The post-harmonization patch embedders consist of convolution, linear, and normalization layers, similar in design to BENDR’s convolutional blocks and \cite{BENDR}. }
	\label{fig:MENDREncoder}
\end{figure}

\begin{figure}
    \centering
	\includegraphics[width=0.5\textwidth]{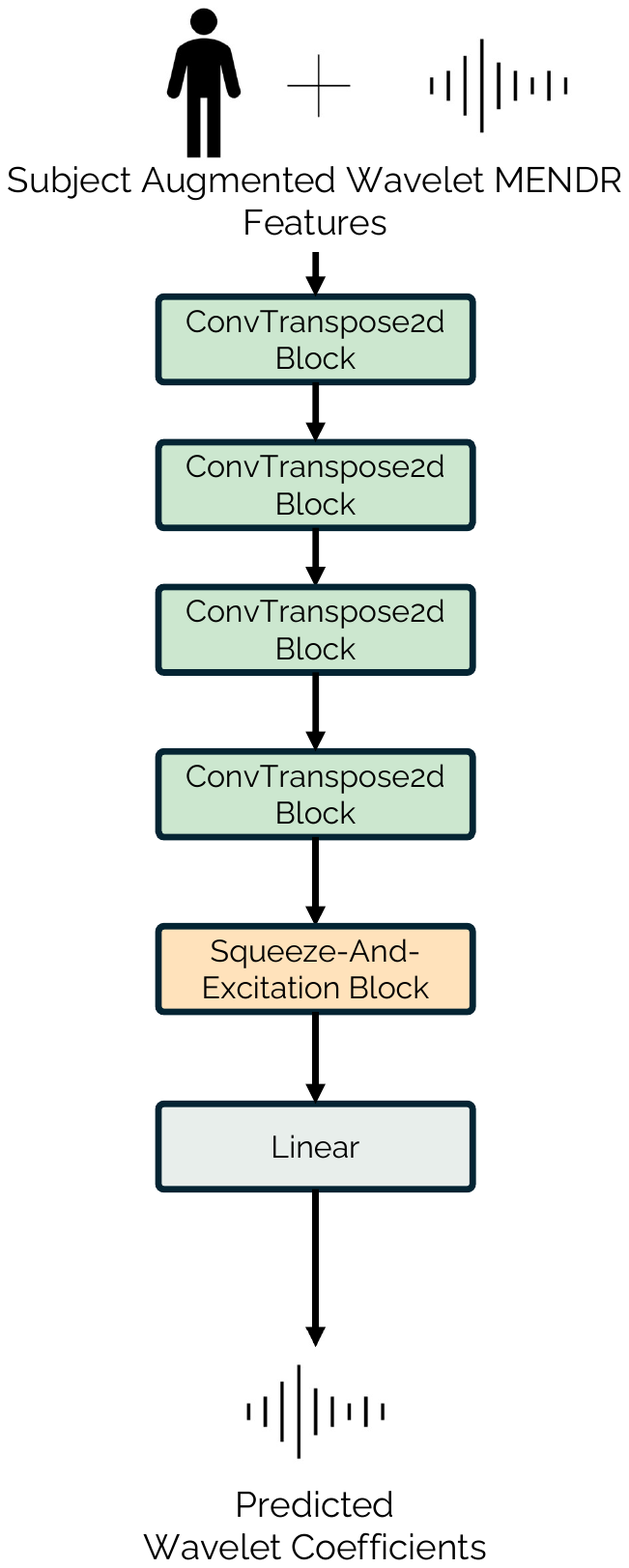}
	\caption[MENDR Decoder Architecture]{In contrast to the encoder, the decoder architecture uses ConvTranspose blocks with GroupNorms, a Squeeze and Excitation block, and a linear layer. Note that before feeding the wavelet features into the decoder, learned subject embeddings are added to the original wavelet embeddings to create subject-augmented wavelet embeddings.}
	\label{fig:MENDRDecoder}
\end{figure}

\begin{figure}
    \centering
	\includegraphics[width=0.5\textwidth]{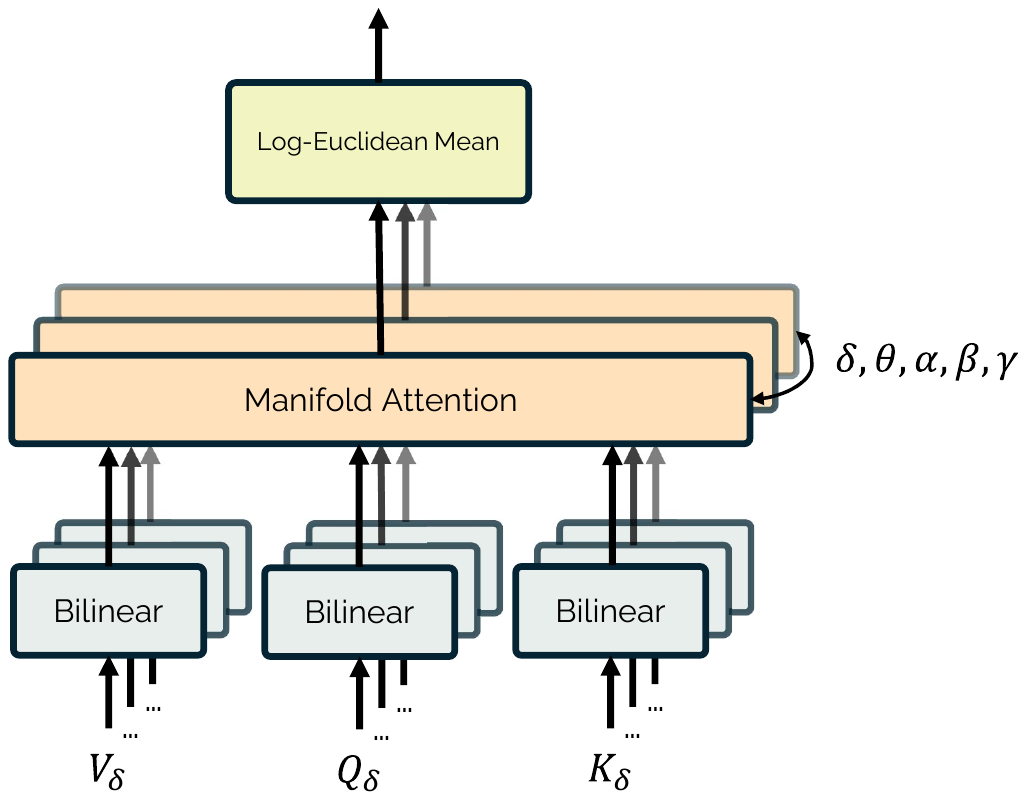}
	\caption[MAtt Architecture.]{The Manifold Attention (MAtt) module.}
	\label{fig:MAtt Architecture}
\end{figure}

\begin{figure*}
\centering
\begin{subfigure}{.5\textwidth}
    \centering \captionsetup{width=.7\linewidth}%
    \includegraphics[width=\linewidth]{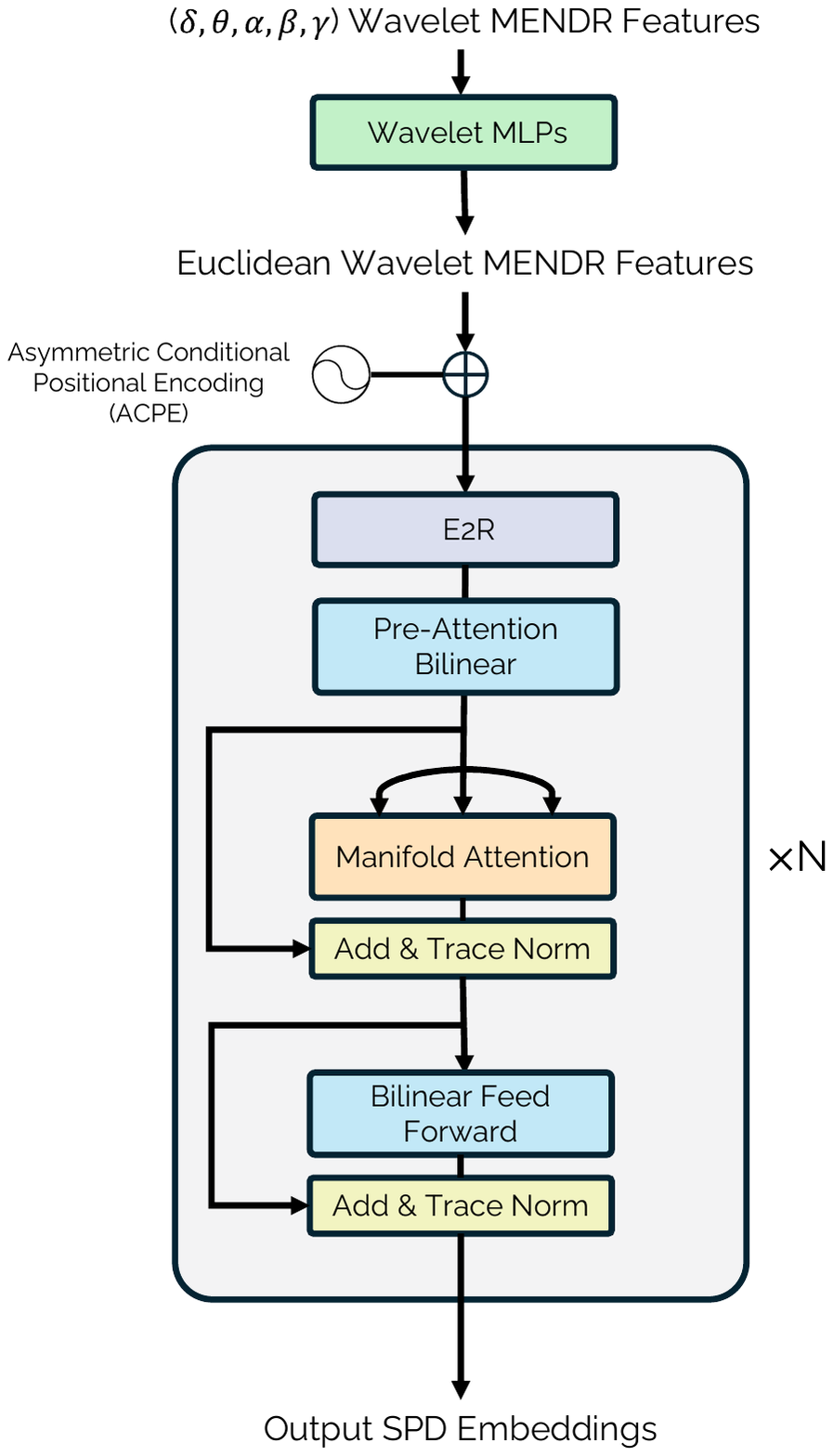}
	\caption[MENDR Tiny Contextualizer]{The MENDR tiny contextualizer architecture. Note that the $||$ symbol in the circle denotes the concatenation operation.}
	\label{fig:MENDRContextualizerTiny}
\end{subfigure}%
\begin{subfigure}{.5\textwidth}
    \centering \captionsetup{width=.7\linewidth}%
	\includegraphics[width=\linewidth]{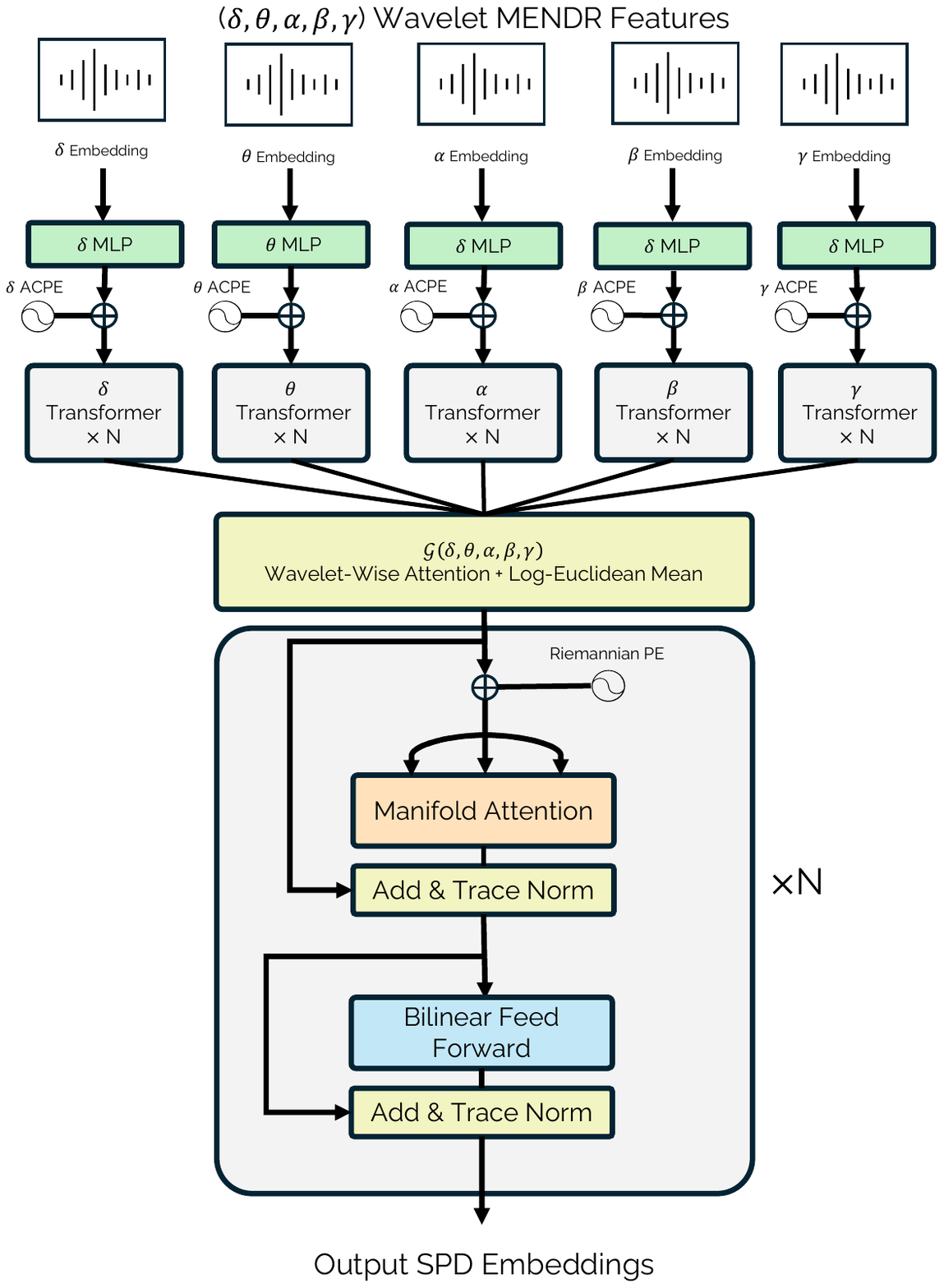}
	\caption[MENDR Large Contextualizer]{The MENDR large contextualizer architecture. Note that each wavelet band transformer is processed in parallel.}
	\label{fig:MENDRContextualizerLarge}
\end{subfigure}
\caption{Different MENDR Contextualizer architectures.}
\end{figure*}

\section{Training Hyperparameters}

Pretraining hyperparameters are listed in Table \ref{table:PretrainHyperparameters}. Hyperparameters for TUAB and TUEV are listed in \ref{table:TUABHyperparameters} and \ref{table:TUEVHyperparameters} respectively. Note that for TUAB and TUEV the scheduler steps at each batch rather than each epoch. Moreover, for the TUEV dataset, due to the class imbalance, we used a weighted random sampler in the training dataset to ensure that the model does not overfit on the training distribution. Finally, only the contextualizer was further finetuned for the downstream tasks -- the autoencoder weights were frozen. The other downstream datasets has identical hyperparameters to either TUAB or TUEV.

\begin{table*}[ht]
  \caption{Wavelet Patch Lengths}
  \label{table:wavelet-patch-lengths}
  \centering
  \begin{tabular}{llll}
    \toprule
    & \multicolumn{2}{c}{128 EEG Signal Hz}  \\
    \cmidrule(r){2-3} 
    Frequency Band & 2 second Patch Length & Brain States \\
    \midrule
    $\delta$ (0-4 Hz) & 8 & Sleep \\
    $\theta$ (4-8 Hz) & 8 & Deep relaxation, inwardly focused \\
    $\alpha$ (8-16 Hz) & 16 & Very relaxed, Passively Attentive \\
    $\beta$  (16-32 Hz) & 32 & Busy, Active Mind, Externally Attentive \\
    $\gamma$ (32-64 Hz) & 64 & Problem Solving, Concentration\\
    $\textbf{high}$ (64-128 Hz) & 128 & N/A \\
    \bottomrule
  \end{tabular}
\end{table*}

\begin{table}
  \caption{Pretraining Hyperparameters}
  \label{table:PretrainHyperparameters}
  \centering
  \resizebox{\linewidth}{!}{%
  \begin{tabular}{ll}
    \toprule
    \textbf{Hyperparameters} & \textbf{Values}  \\
    \midrule
    Train Split & 0.9 \\
    Val Split & 0.1 \\
    Learning Rate & 0.001 \\
    $L_{2}$ weight decay & 0.0005 \\
    Batch Size & 256 \\
    Autoencoder Training Epochs & 30 \\
    Tiny Contextualizer Training Epochs & 10 \\
    Large Wavelet Contextualizer Training Epochs & 5 \\
    Large Combined Contextualizer Training Epochs & 5 \\
    \$ of Tiny Contextualizer Transformer Layers & 6 \\
    \# of Wavelet Contextualizer Transformer Layers & 6 \\
    \# of Large Combined Contextualizer Transformer Layers & 8 \\
    Tiny Contextualizer Matrix Embedding Dimension & 19 \\
    Large Contextualizer Matrix Embedding Dimension & 6 \\
    Initial Temperature & 1.0 \\
    Mask Ratio & 0.2 \\
    Gradient Clip Value & $1 * 10^{7}$ \\
    \# of negatives in LOO & $32$ \\
    Optimizer & AdamW \\
    Scheduler & CosineAnnealingLR \\
    Scheduler $\eta$ min & $1 * 10^{-7}$ \\
    Scheduler $T_{max}$ & \# of training epochs \\
    \bottomrule
  \end{tabular}}
\end{table}

\begin{table}
  \caption{TUAB Hyperparameters}
  \label{table:TUABHyperparameters}
  \centering
  \resizebox{\linewidth}{!}{%
  \begin{tabular}{ll}
    \toprule
    \textbf{Hyperparameters} & \textbf{Values}  \\
    \midrule
    Learning Rate & 0.001 \\
    $L_{2}$ weight decay & 0.01 \\
    Batch Size & 256 \\
    Training Epochs & 3 \\
    Gradient Clip Value & $1 * 10^{8}$ \\
    Optimizer & AdamW \\
    Scheduler & CosineAnnealingLR \\
    Scheduler $\eta$ min & $1 * 10^{-7}$ \\
    Scheduler $T_{max}$ & \# of training epochs * \$ of batches per epoch \\
    \bottomrule
  \end{tabular}}
\end{table}

\begin{table}
  \caption{TUEV Hyperparameters}
  \label{table:TUEVHyperparameters}
  \centering
  \resizebox{\linewidth}{!}{%
  \begin{tabular}{ll}
    \toprule
    \textbf{Hyperparameters} & \textbf{Values}  \\
    \midrule
    Learning Rate & 0.001 \\
    $L_{2}$ weight decay & 0.1 \\
    Batch Size & 256 \\
    Training Epochs & 10 \\
    Gradient Clip Value & $1 * 10^{8}$ \\
    CrossEntropyLoss Label Smoothing & 0.1 \\
    Optimizer & AdamW \\
    Scheduler & CosineAnnealingLR \\
    Scheduler $\eta$ min & $1 * 10^{-7}$ \\
    Scheduler $T_{max}$ & \# of training epochs * \$ of batches per epoch \\
    \bottomrule
  \end{tabular}}
\end{table}

\section{Visualization of the SPD Patch Embeddings as Ellipsoids} 

\begin{figure*}[ht]
\centering

\begin{subfigure}[b]{\textwidth}
  \includegraphics[width=\linewidth]{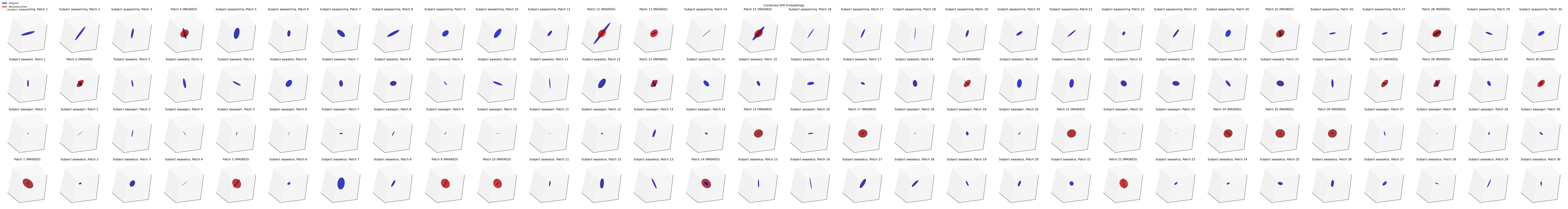}
  \caption{Riemannian SPD Matrix Embeddings in the first pretraining epoch of the tiny contextualizer.}
  \label{fig:Epoch1Pretraining} 
\end{subfigure}
\medskip 
\begin{subfigure}[b]{\textwidth}
  \includegraphics[width=\linewidth]{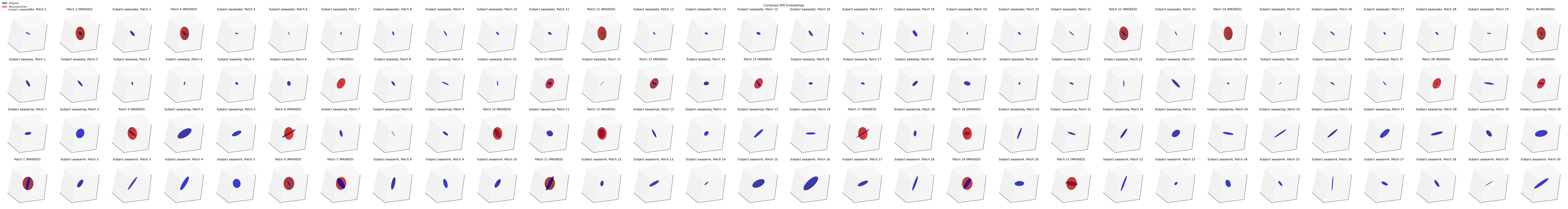}
  \caption{Riemannian SPD Matrix Embeddings in the last pretraining epoch of the tiny contextualizer.}
  \label{fig:EpochLastPretraining}
\end{subfigure}

\caption[Pretraining Visualization]{ Visualizations of matrix embeddings of a single 30-second segment during pre-training of the tiny contextualizer. Figure \ref{fig:Epoch1Pretraining} represents the learned embeddings during the first pretraining epoch, and \ref{fig:EpochLastPretraining} represents the embeddings in the last training epoch.}

\end{figure*}

Since our SPD embeddings are derived from the covariance of each patch, we can think of each patch as a miniature dataset and perform more nuanced statistical analysis of these embeddings compared to typical vector embeddings. Consider an SPD matrix embedding $\mathbf{A} \in \mathcal{M}_{SPD}$. Since $\mathbf{A}$ is real and symmetric by construction, the factorization of $\mathbf{A}$ is always possible: 

\begin{equation}
    \mathbf{A} = PDP^{-1} = PDP^{T}
    \label{eq:Eigendecomposition}
\end{equation}

where $P$ is a matrix with the orthogonal eigenvectors of $\mathbf{A}$ and $D$ is the diagonal matrix that contains the eigenvalues of $\mathbf{A}$. Furthermore, an ellipsoid centered at the origin is defined by the solutions $\mathbf{x}$ to the equation:

\begin{equation}
    \mathbf{x}^{T}M\mathbf{x} = \mathbf{1}
    \label{eq:EllipseEquation}
\end{equation}

where $M$ must be a positive definite matrix. To align equation \eqref{eq:Eigendecomposition} with \eqref{eq:EllipseEquation}, we find all solutions $\mathbf{x}$ such that $\mathbf{x}^{T}PDP\mathbf{x} = 1$ because note that: 

\begin{equation}
    \mathbf{x}^{T}PDP\mathbf{x} = (P^{T}\mathbf{x})^{T} D (P^{T}\mathbf{x}) = \mathbf{y}^{T}D\mathbf{y}=1
    \label{eq:SPDEllipsoid}
\end{equation}

if we let $\mathbf{y} = P^{T}\mathbf{x}$. Thus, $\mathbf{A}$ can be represented as an ellipse where the transformation matrix $M$ is a diagonal matrix $D$ with eigenvalues of $\mathbf{A}$ in its entries. Note that if all eigenvalues are the same, the ellipse would be a circle; however, the eigendecomposition would not be differentiable in this case. The orthogonal eigenvectors in $P$ essentially represent a pure rotation of the canonical basis axes $e_{1}, \ldots, e_{n}$ so we are essentially rotating an $N$-dimensional circle with radius $1$ along all axes, and the axes are aligned with the canonical basis axes. The quadratic form of equation \ref{eq:SPDEllipsoid} when converted to the canonical form of the ellipsoid is:

\begin{equation}
    \mathbf{y}^{T}D\mathbf{y} = \lambda_{1}y_{1}^{2} + \ldots + \lambda_{n}y_{n}^{2} = \frac{y_{1}^{2}}{c_{1}^{2}} + \ldots + \frac{y_{1}^{2}}{c_{1}^{2}} = 1
\end{equation}

if we let $\lambda_{i} = \frac{1}{c_{1}^{2}}$. Thus, the length of the ellipsoid’s axes in dimension $i$ is $2\sqrt{\frac{1}{\lambda_{i}}}$, which is guaranteed to be a real number because $\lambda_{i} > 0$. Unfortunately, we can’t visualize more than $3$ dimensions in practice. Borrowing the intuition from Principal Component Analysis (PCA), we have already fit an $n$-dimensional ellipsoid to each patch through the construction of covariance matrix embeddings, where each component is a principal axis \cite{pearson_1901_1430636}. Moreover, we have already calculated the eigenvalue decompositions of each embedding. Then, we need to sort the eigenvalues in decreasing order and take the top $3$ principal components for visualization. Figures \ref{fig:Epoch1Pretraining} and \ref{fig:EpochLastPretraining} represent the visualizations of the embeddings during Tiny contextualizer pretraining, with the blue representing the original embedding and the red representing the predicted embedding from the  contextualizer.

\section{Manifold Explainability via UMAP and Riemannian TSNE}

\begin{figure*}[ht]
\centering
\begin{subfigure}[b]{0.8\textwidth}
  \includegraphics[width=\linewidth]{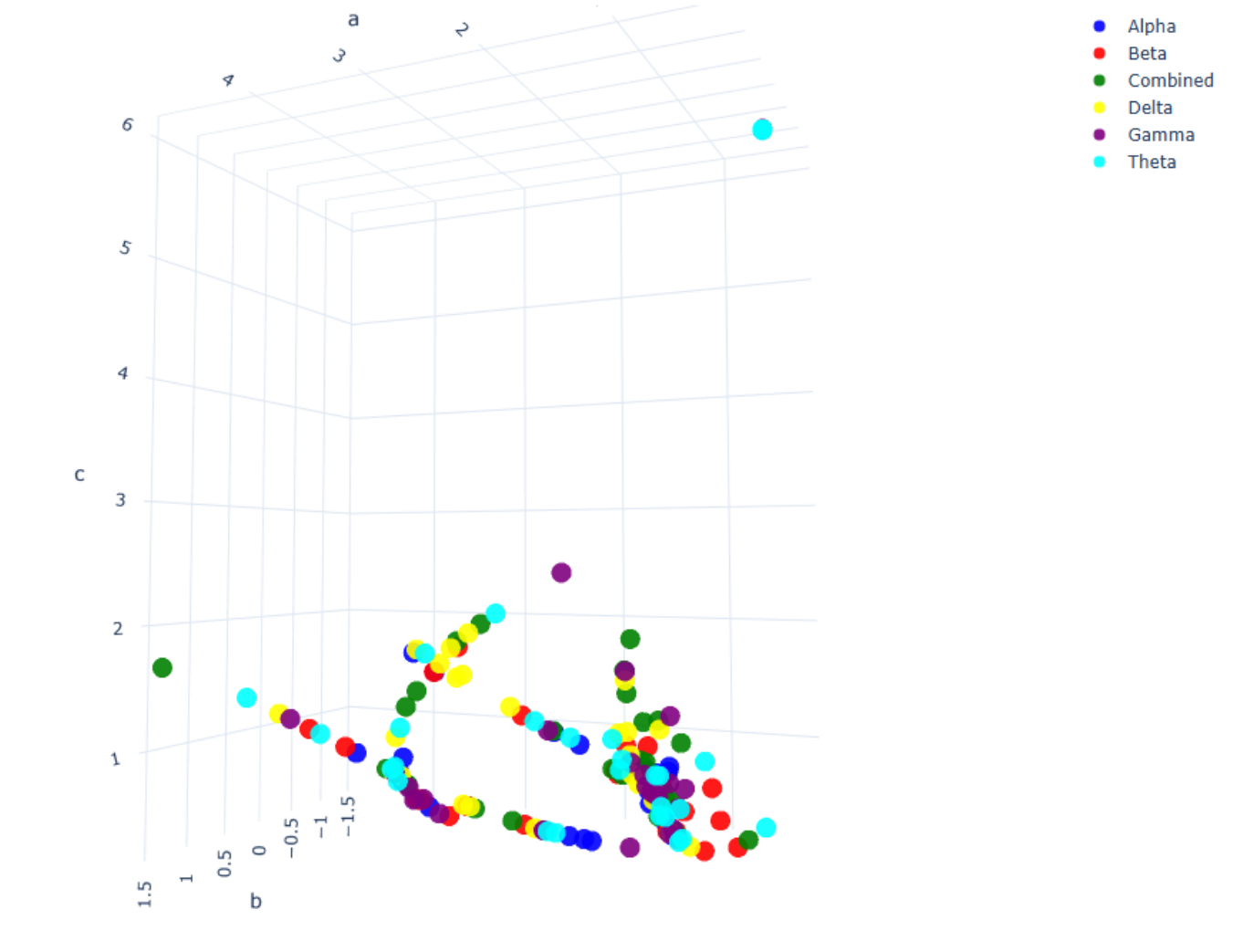}
  \caption{Riemannian-Riemannian TSNE of a single 30-second segment of embeddings from the validation set in first last wavelet contextualizer pretraining epoch.}
  \label{fig:Epoch1TSNE} 
\end{subfigure}
\medskip 
\begin{subfigure}[b]{0.8\textwidth}
  \includegraphics[width=\linewidth]{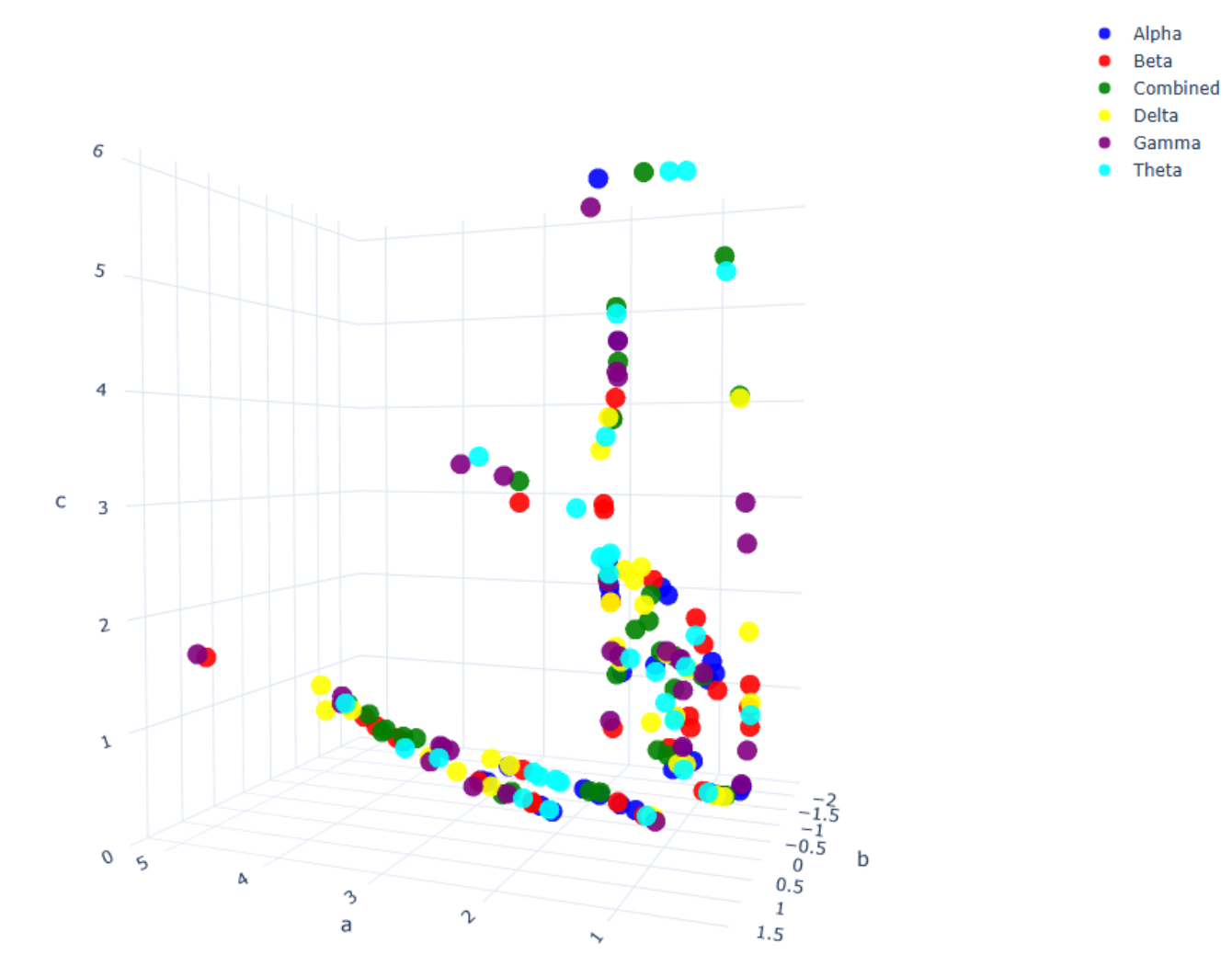}
  \caption{Riemannian-Riemannian TSNE of a single 30-second segment of embeddings from the validation set in the last wavelet contextualizer pretraining epoch.}
  \label{fig:EpochLastTSNE}
\end{subfigure}

\caption[Pretraining Visualization]{ Visualizations of matrix embeddings of a single 30-second segment during pre-training of the tiny contextualizer. Figure \ref{fig:Epoch1Pretraining} represents the learned embeddings during the first pretraining epoch, and \ref{fig:EpochLastPretraining} represents the embeddings in the last training epoch.}

\end{figure*}

In this section, we present more visualizations of the learned embeddings. When pretraining the wavelet contextualizer, to gain visibility into the learned neural manifolds, we utilize Riemannian-Riemannian t-SNE proposed by \cite{de2025geometry} to visualize higher-dimensional SPD matrices in 3 dimensions. Figure \ref{fig:Epoch1TSNE} and \ref{fig:EpochLastTSNE} represent the learned embeddings from the first pretraining to the last pretraining epoch of the wavelet contextualizer in the large MENDR model.

\begin{figure*}
\centering
\begin{subfigure}{.33\textwidth}
  \centering
  \includegraphics[width=\linewidth]{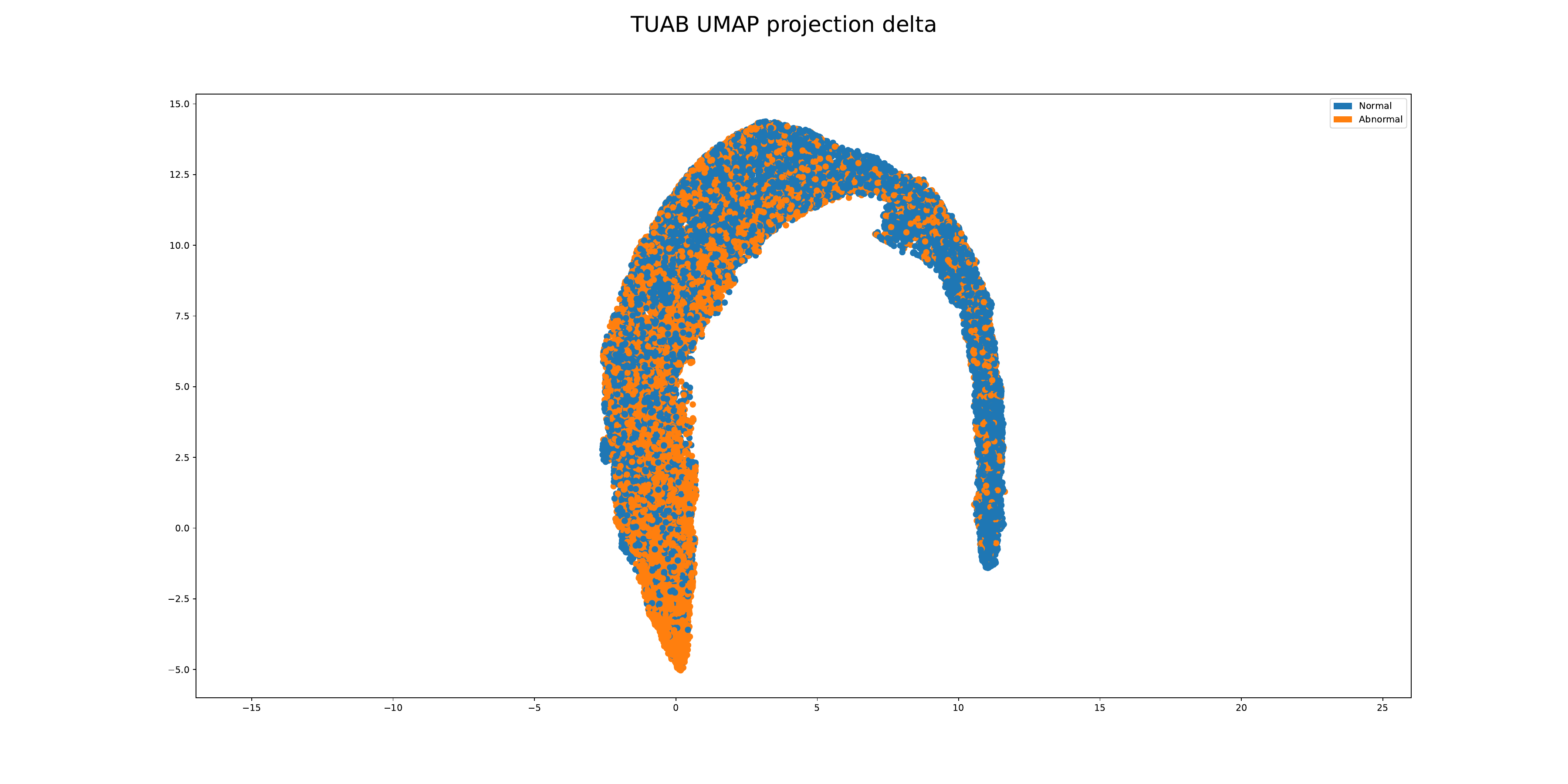}
\end{subfigure} \hfil
\begin{subfigure}{.33\textwidth}
  \centering
  \includegraphics[width=\linewidth]{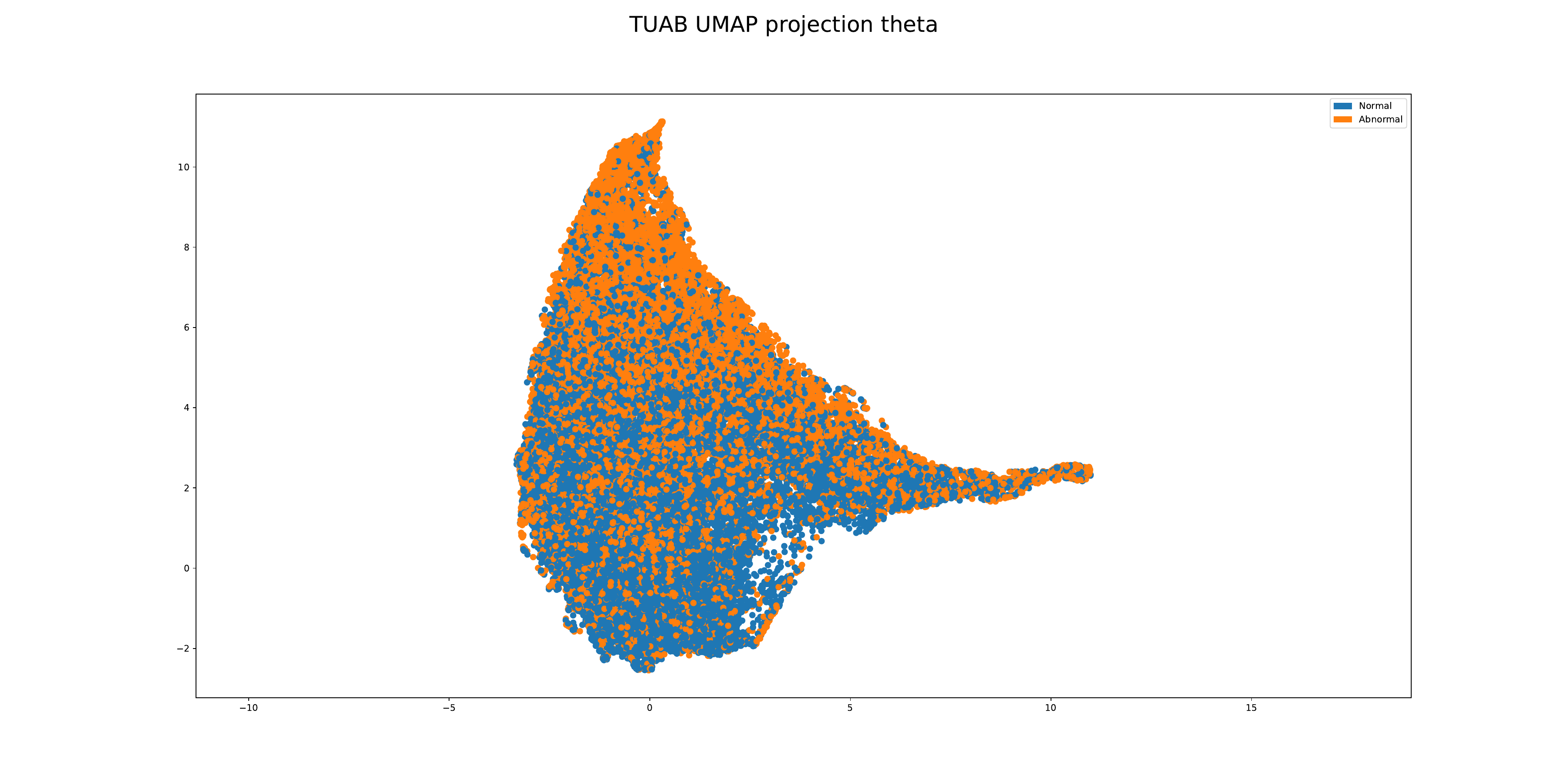}
\end{subfigure} \hfil
\begin{subfigure}{.33\textwidth}
  \centering
  \includegraphics[width=\textwidth]{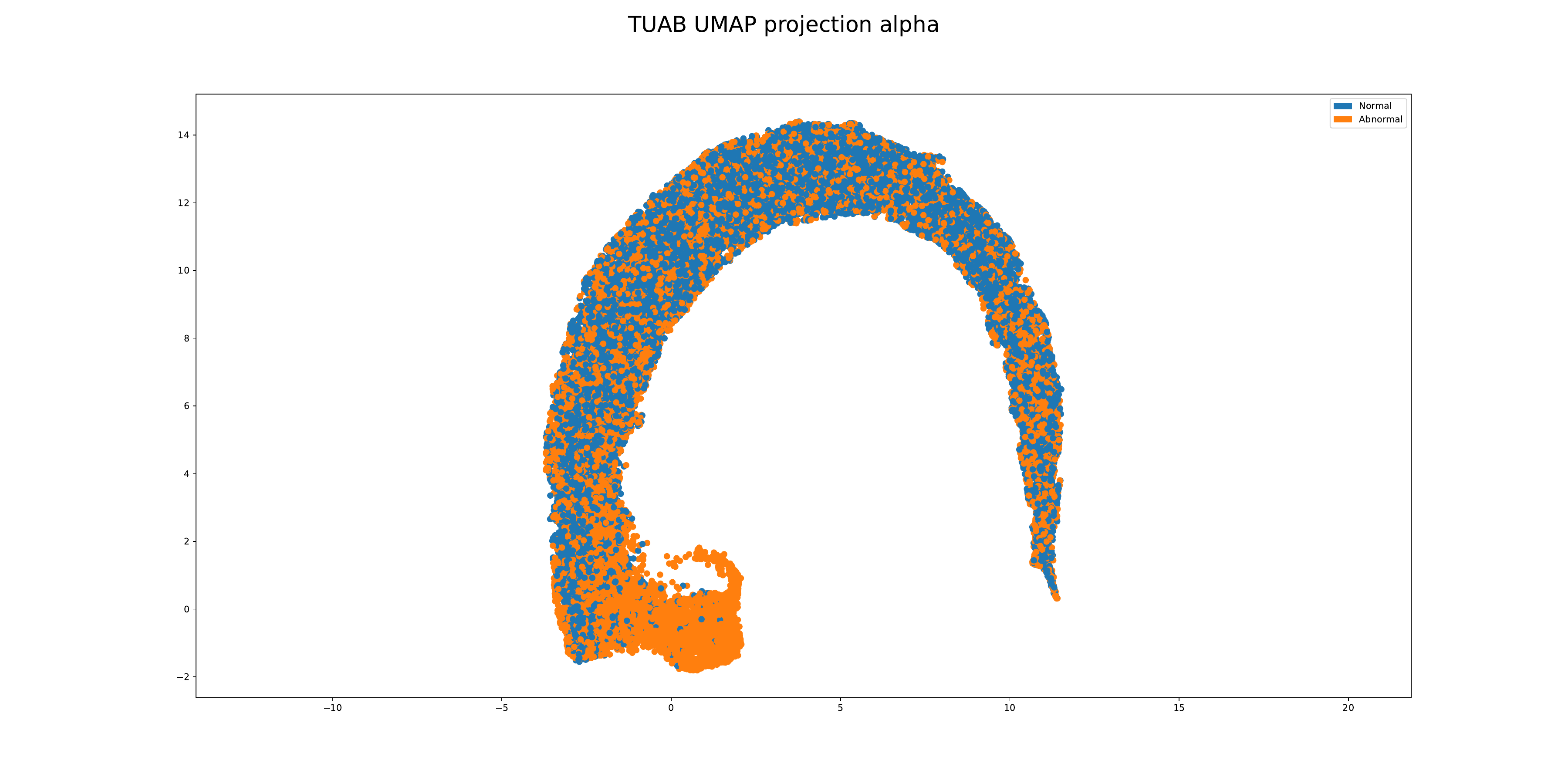}
\end{subfigure}
\medskip
\begin{subfigure}{.33\textwidth}
  \centering
  \includegraphics[width=\linewidth]{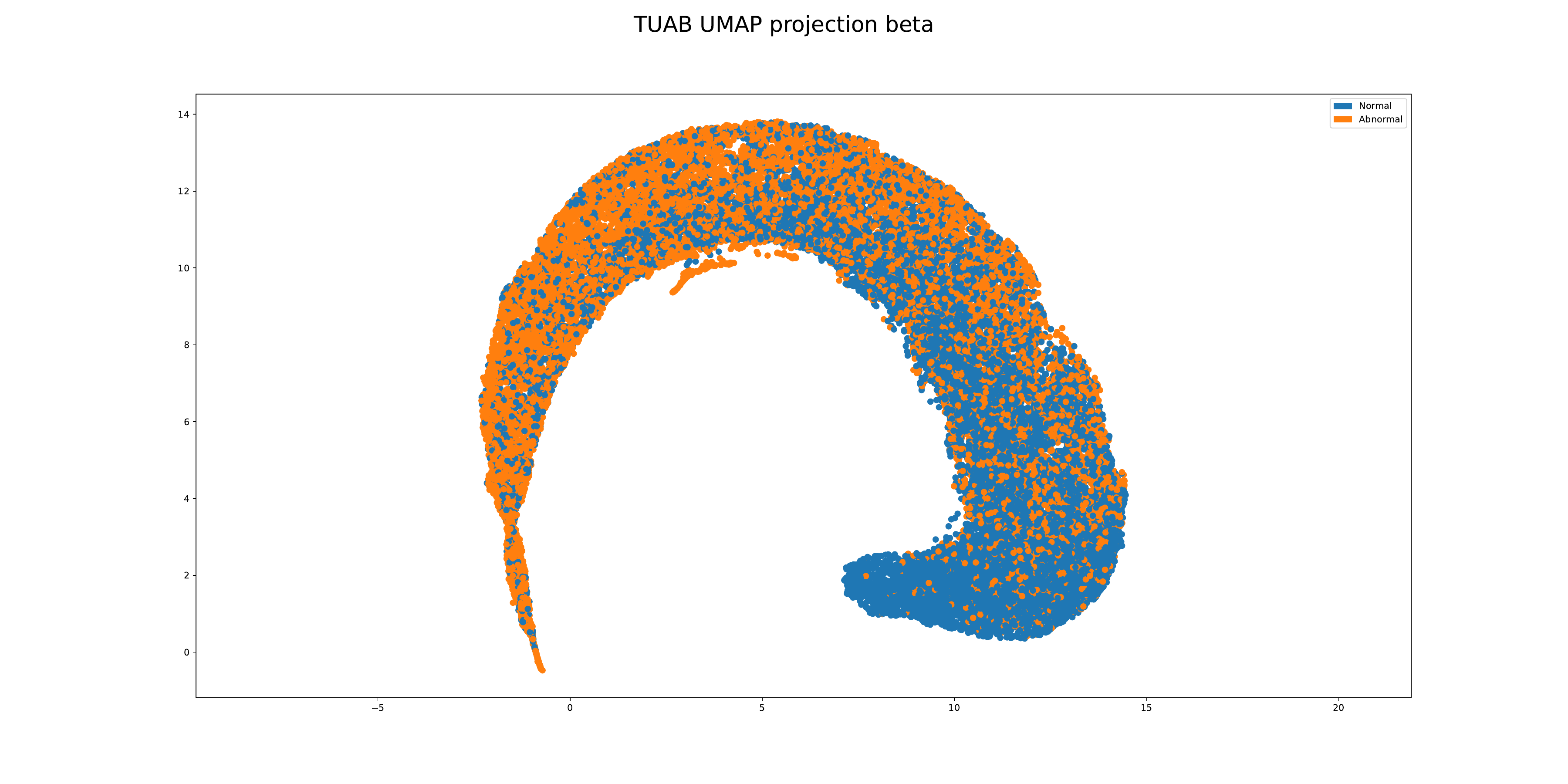}
\end{subfigure} \hfil
\medskip
\begin{subfigure}{.33\textwidth}
  \centering
  \includegraphics[width=\linewidth]{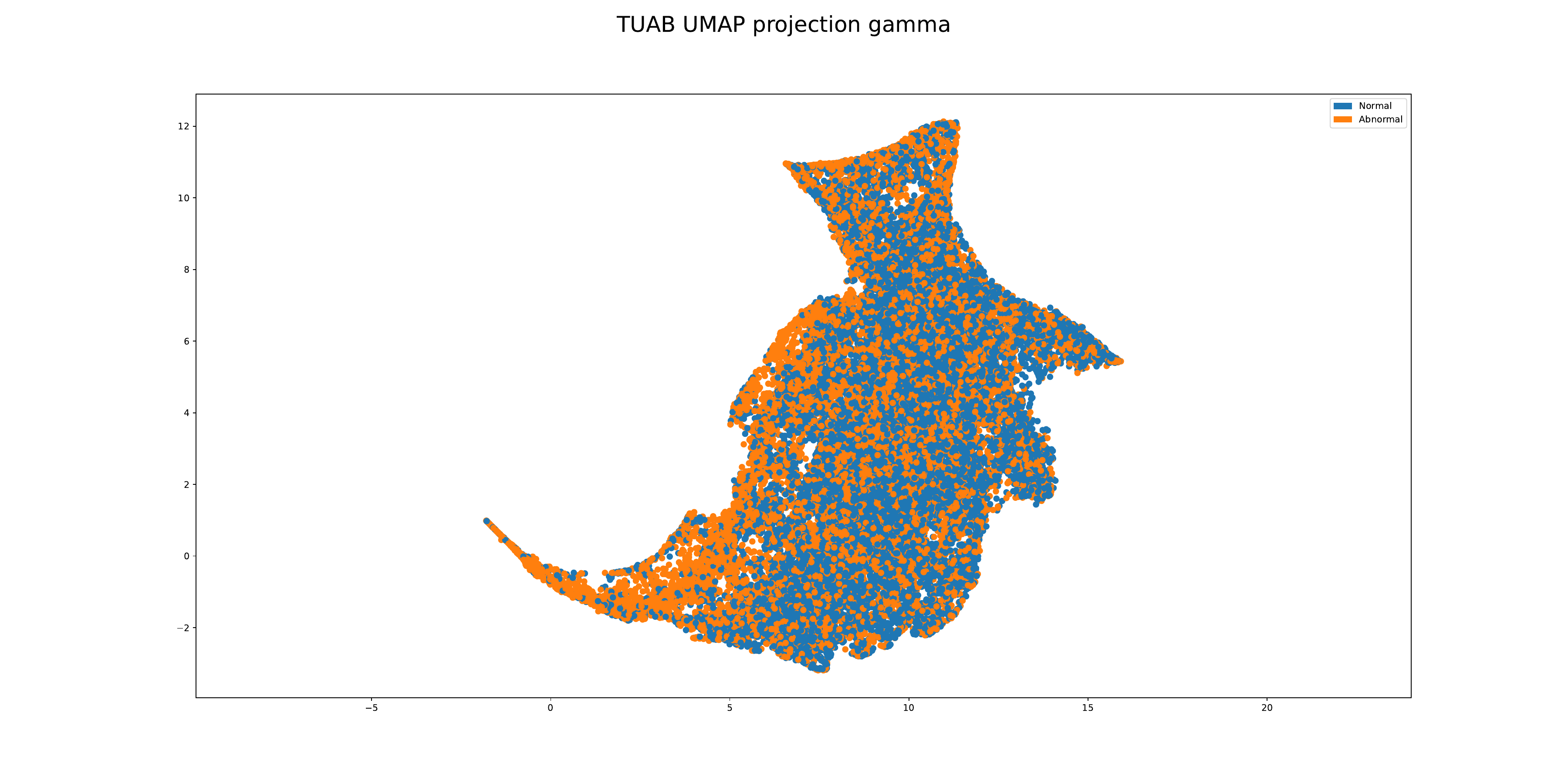}
\end{subfigure} \hfil
\begin{subfigure}{.33\textwidth}
  \centering
  \includegraphics[width=\linewidth]{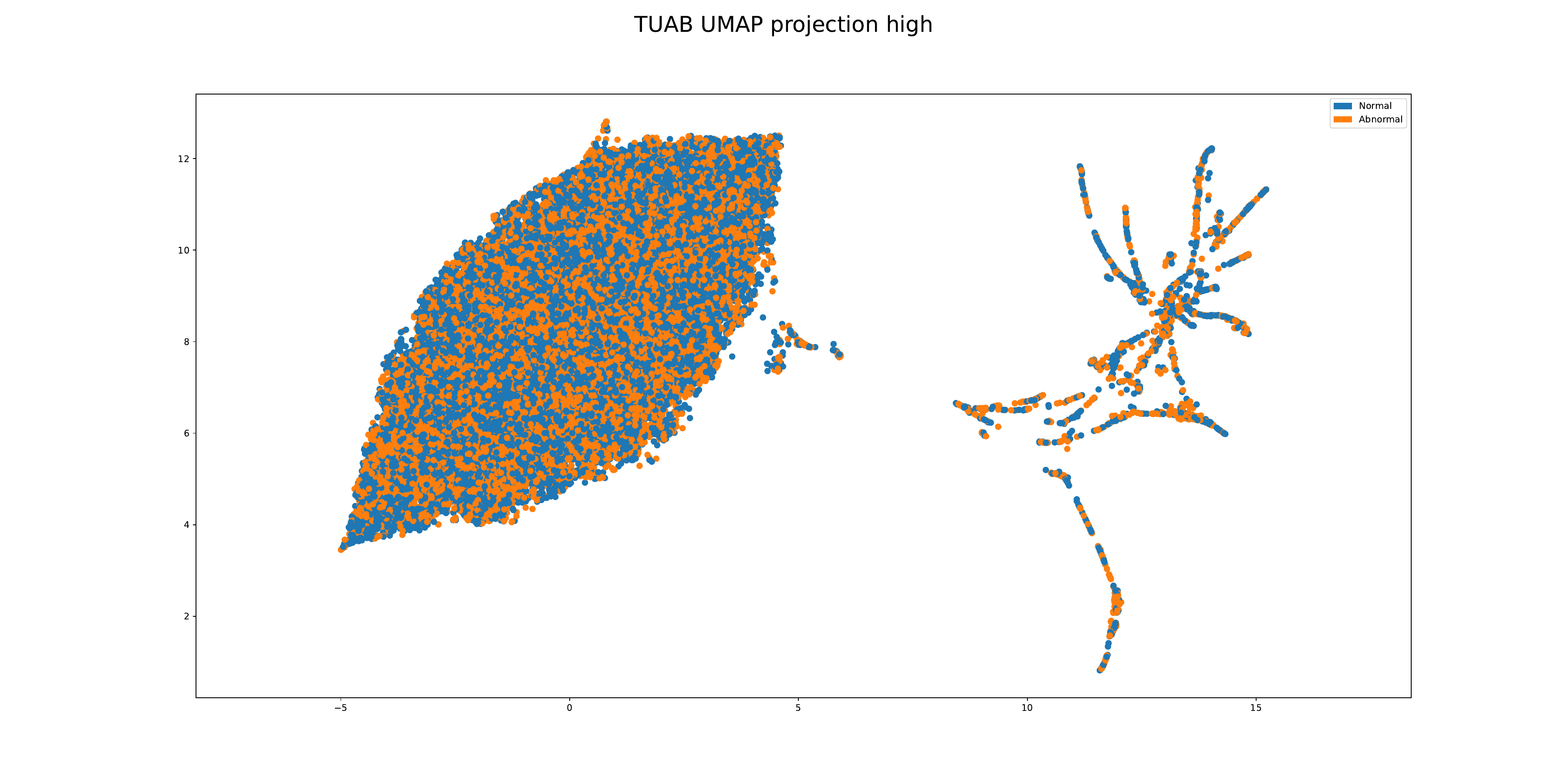}
\end{subfigure}

\caption{TUAB UMAP embeddings of the Wavelet SPD and combined embeddings after calculating their tangent space projection for the Large Contextualizer. Orange points indicate abnormal EEGs, and blue points indicate normal ones. The combined embedding is the only embedding used for prediction.}
\label{fig:TUABWaveletManifoldUMAPLarge}
\end{figure*}

\begin{figure*}
\centering
\begin{subfigure}{.33\textwidth}
  \centering
  \includegraphics[width=\linewidth]{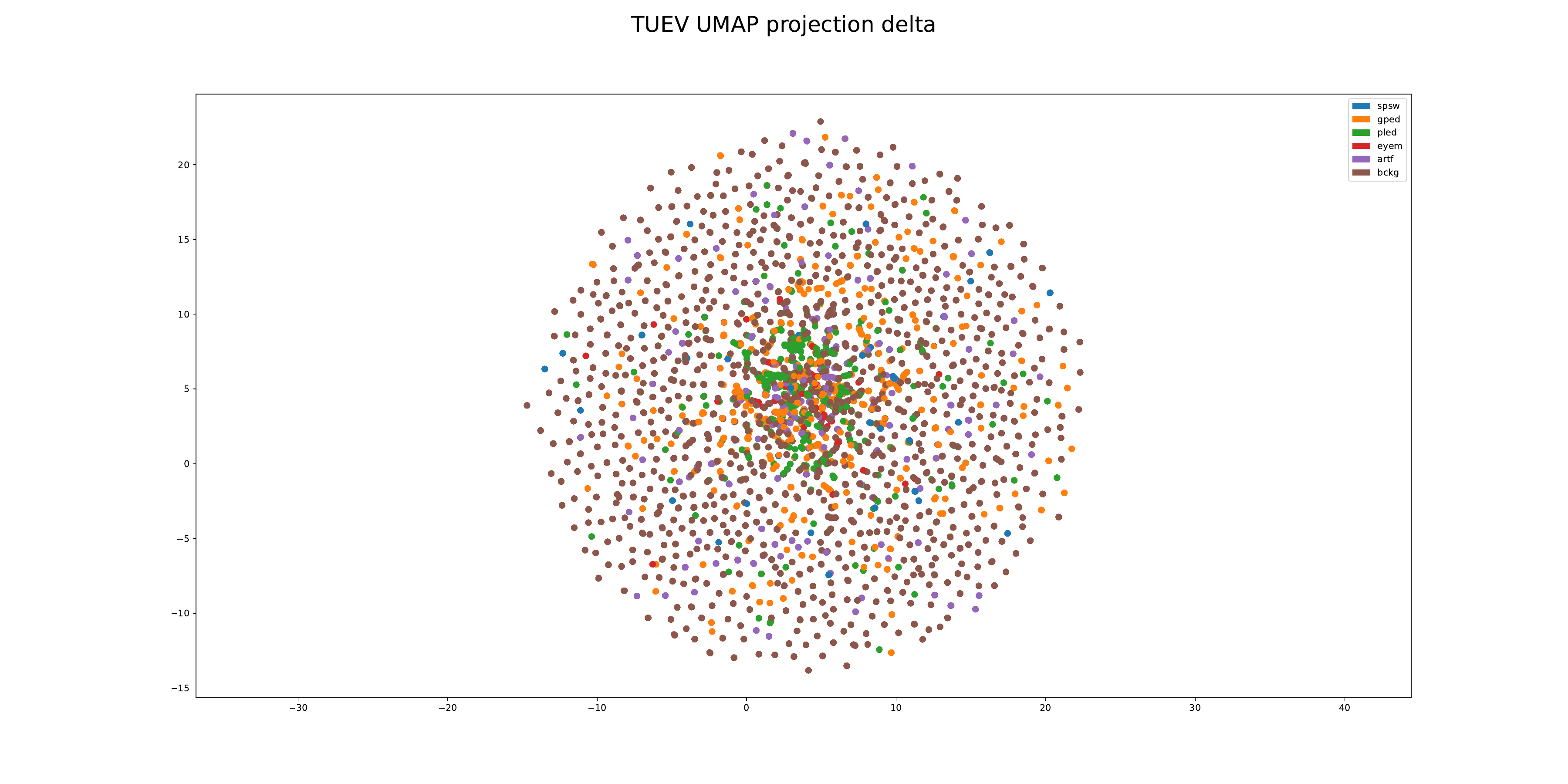}
\end{subfigure} \hfil
\begin{subfigure}{.33\textwidth}
  \centering
  \includegraphics[width=\linewidth]{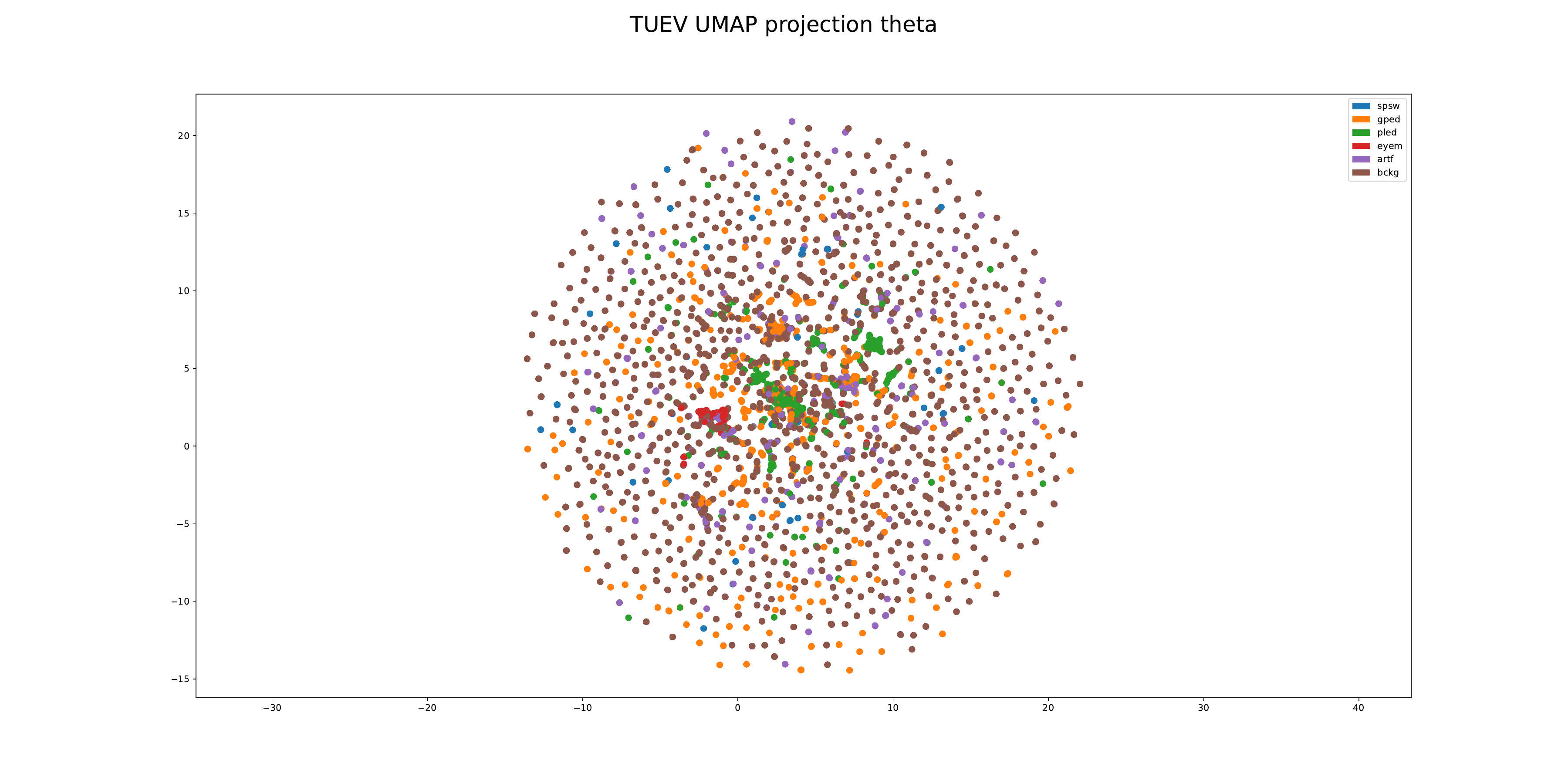}
\end{subfigure} \hfil
\begin{subfigure}{.33\textwidth}
  \centering
  \includegraphics[width=\textwidth]{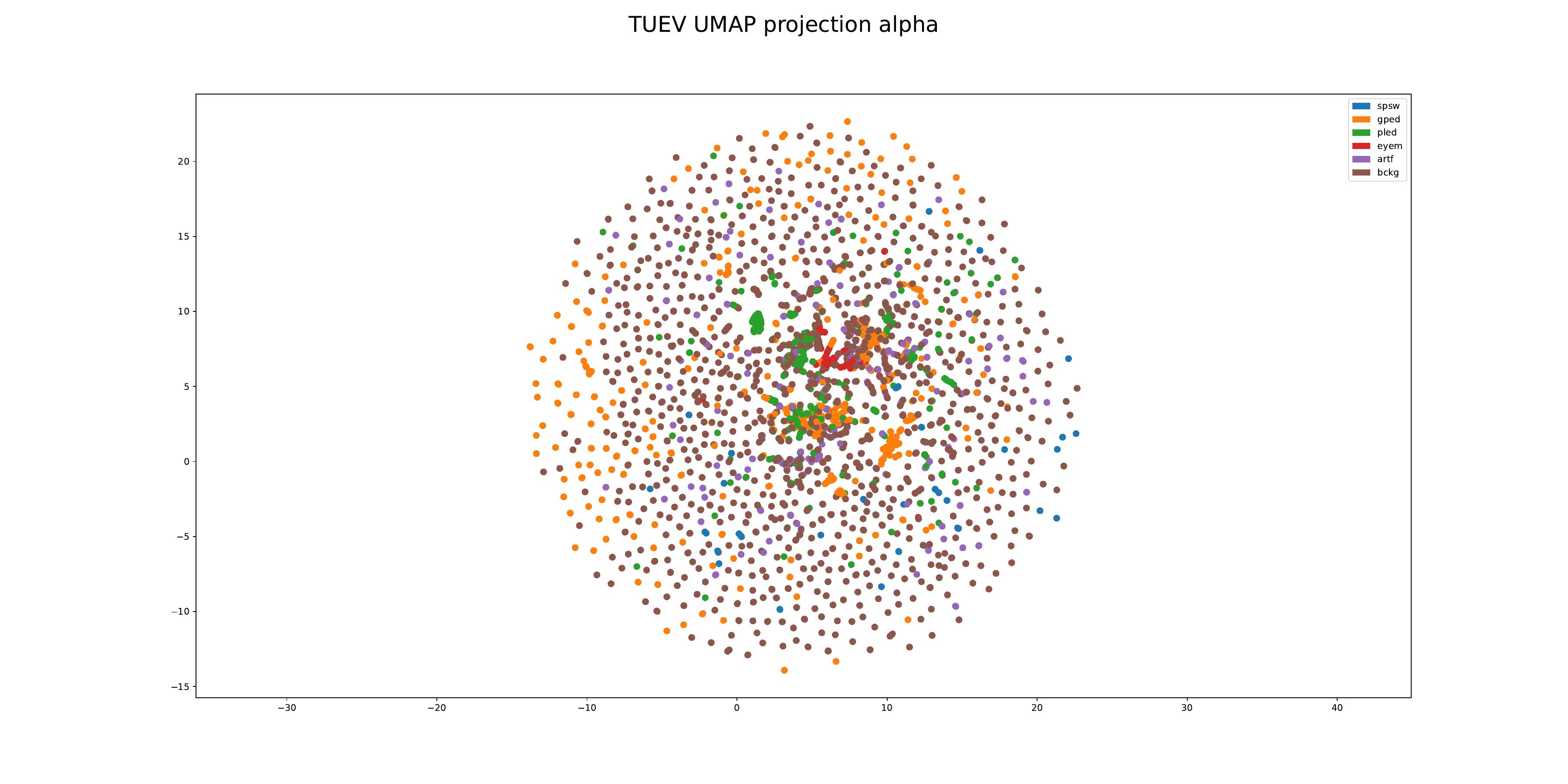}
\end{subfigure}
\medskip
\begin{subfigure}{.33\textwidth}
  \centering
  \includegraphics[width=\linewidth]{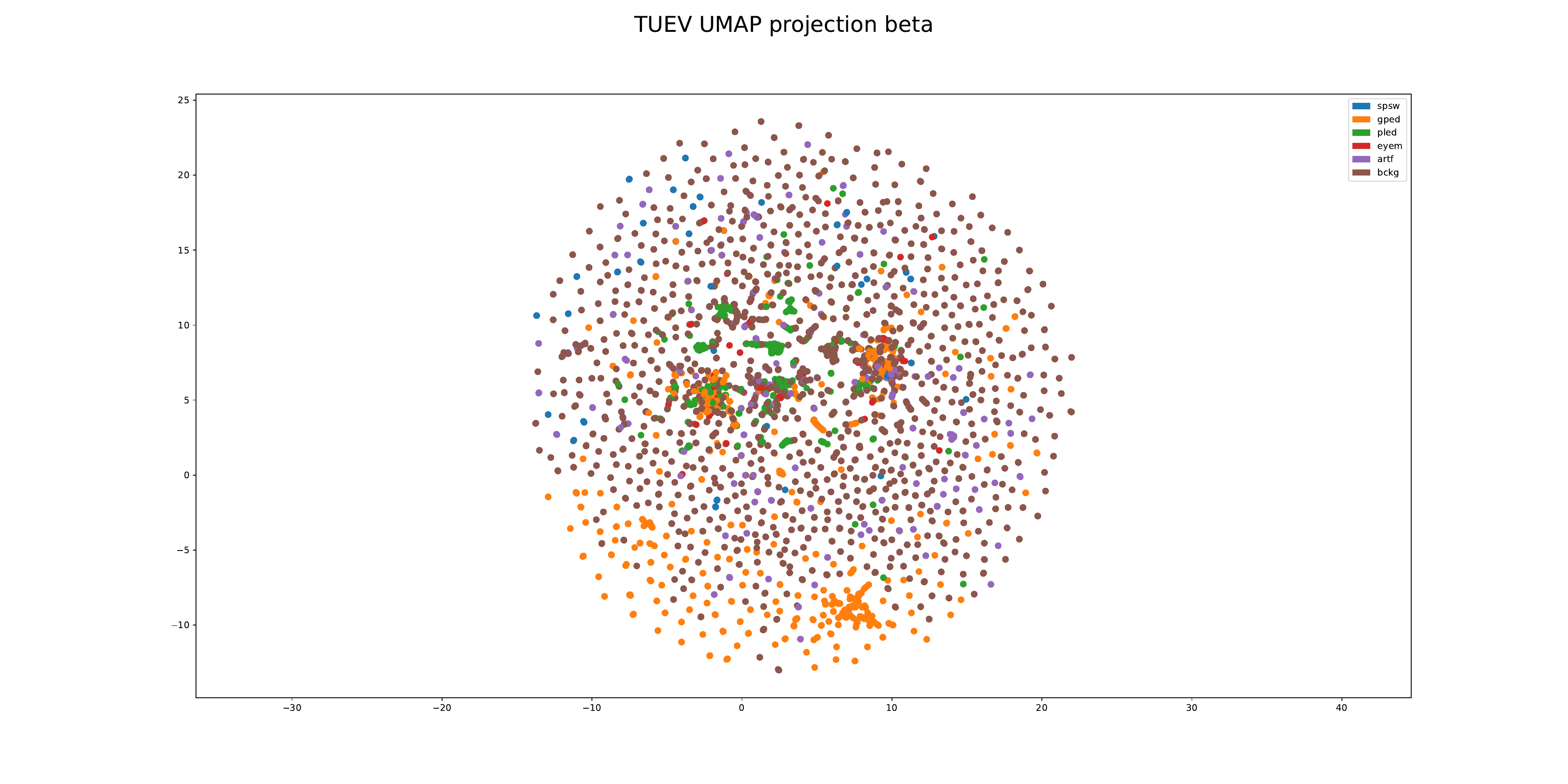}
\end{subfigure} \hfil
\medskip
\begin{subfigure}{.33\textwidth}
  \centering
  \includegraphics[width=\linewidth]{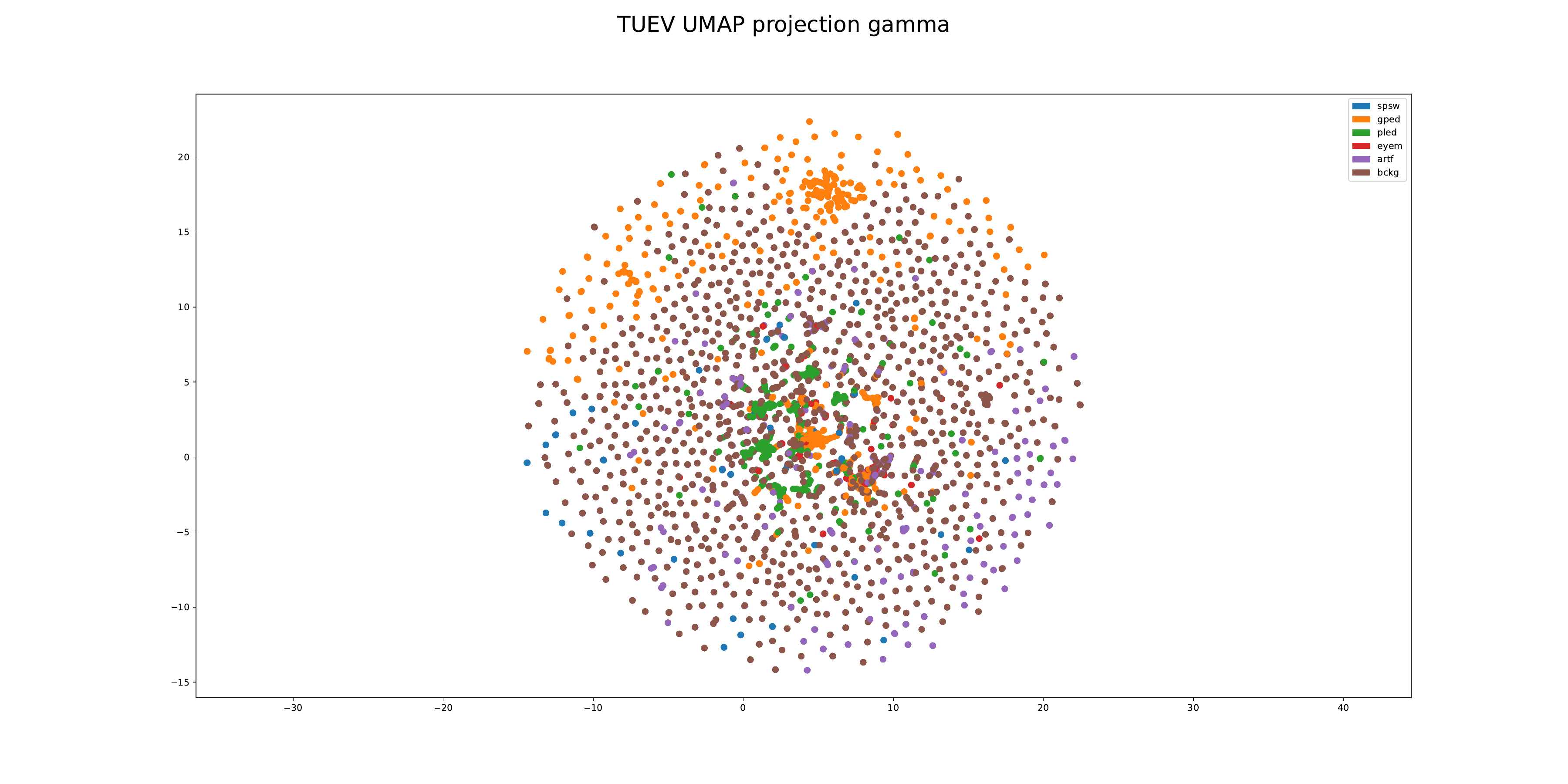}
\end{subfigure} \hfil
\begin{subfigure}{.33\textwidth}
  \centering
  \includegraphics[width=\linewidth]{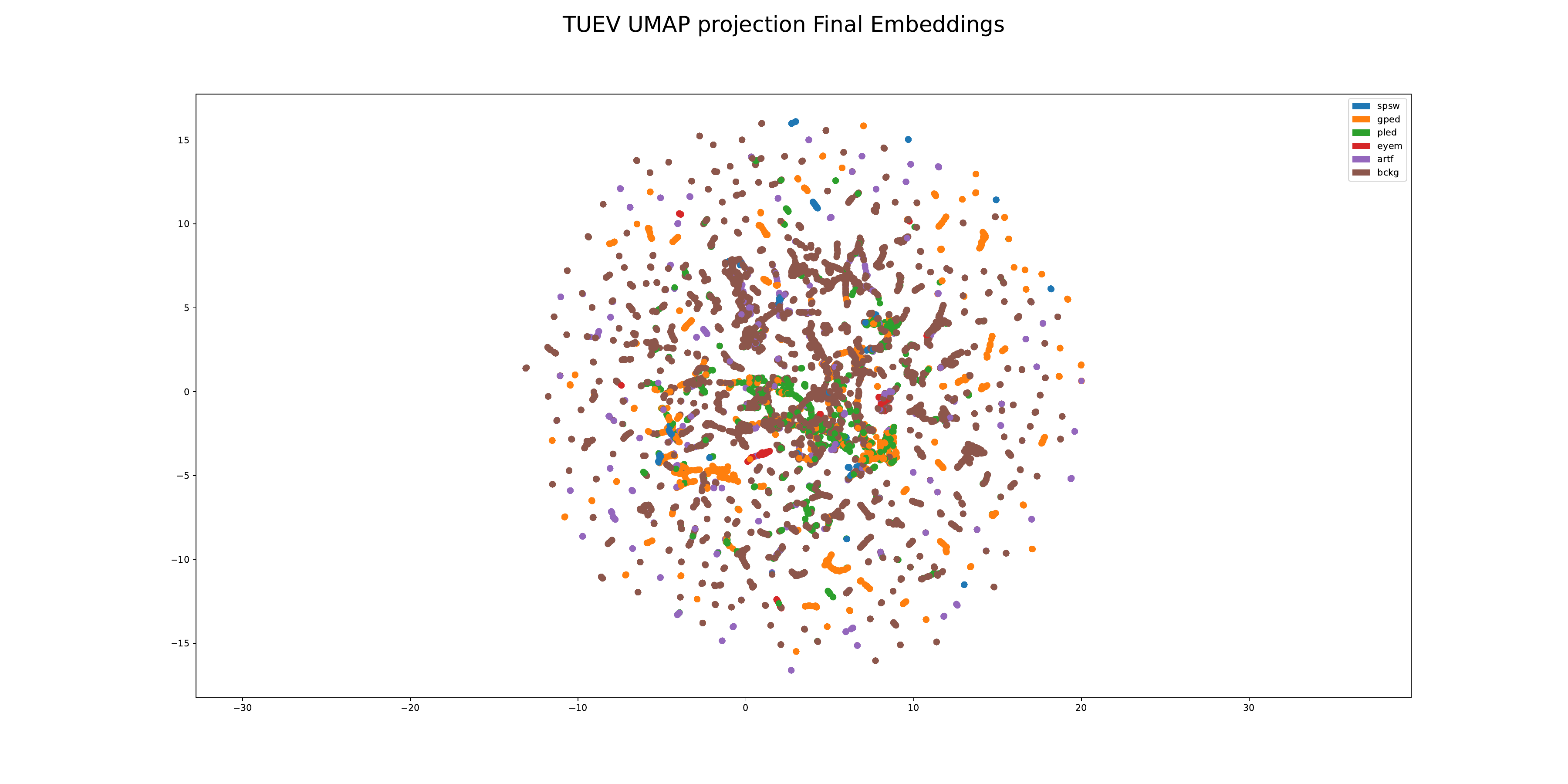}
\end{subfigure}

\caption{TUEV UMAP embeddings of the Wavelet SPD and combined embeddings after calculating their tangent space projection for the Large Contextualizer. Different colored classes are denoted in the legend.}
\label{fig:TUEVWaveletManifoldUMAPLarge}
\end{figure*}

For the two main downstream classification datasets, TUAB and TUEV we visualize the matrix embeddings of each frequency band and the final embeddings in the large contextualizer before being passed into the final decoder. This allows us to explain which frequency bands contribute most to the separability of classes by visualizing the learned manifolds. Since the UMAP for the combined embeddings of TUAB was already included in the original paper, we include a MENDR Contextualizer Large with the High embeddings in Figure \ref{fig:TUABWaveletManifoldUMAPLarge}. For TUEV, we include the final embeddings and disregard the high embeddings, since the model performed with them, in Figure \ref{fig:TUEVWaveletManifoldUMAPLarge}.

\section{Additional Downstream Datasets}

We present the two other downstream datasets here. In general, MENDR matches baselines, but does not outperform LaBraM, which is in part due to the size of the LaBraM models. 

\subsection{MoBI Dataset}

Similar to LaBraM \cite{LABRAM}, we also evaluate the MENDR Tiny and Large models, without high frequencies, on the MoBI dataset \cite{he2018mobile}. The MoBI dataset comprises EEG data collected from eight subjects walking on a treadmill, along with six goniometers positioned on the subjects to record their bilateral joint angles at the hip, knee, and ankle. We only regress on one leg, giving us six targets, and each subject’s recordings were 15 minutes long. For each subject, we consider the first 10 minutes for the training dataset and the last 5 minutes for evaluation. Again, the dataset was truncated to the standard 19 channels, resampled to 128 Hz, and divided into 6-second segments. Like LaBraM \cite{LABRAM}, since MoBI is a regression task, we use \textbf{Pearson’s correlation}, the $R^{2}$ coefficient of determination, regression score, and Root Mean Squared Error (RMSE) as our evaluation metrics. We also optimize via the Mean Squared Error (MSE) loss. Table \ref{tbl:MOBIResultTable} presents our results, including results from the original LaBraM paper and BIOT paper \cite{LABRAM, BIOT} for comparison. 

\begin{table}[ht]
    \centering
    \caption{Results on MOBI Dataset}
    \label{tbl:MOBIResultTable}
    \resizebox{\linewidth}{!}{%
    \begin{tabular}{ccccc}
        \hline
        \hline
        & & \multicolumn{3}{c}{MOBI (Multi-variate Regression)}\\
        \cmidrule(r){3-5}
        Methods & Model Size & Pearson’s Correlation & R2 Score & RMSE $\downarrow$ \\
        \hline
        \textbf{MENDR-Tiny} & 1.815M &  $0.4962 \pm 0.0132$ & $0.2428 \pm 0.0032$ & $0.1458 \pm 0.0002$ \\
        \textbf{MENDR-Large} & 1.817M & $0.5024 \pm 0.0033$ & $0.2433 \pm 0.0036$ & $0.1455 \pm 0.0003$ \\
        \hline
        SPaRCNet  & 0.79M & $0.4561 \pm 0.0161$ & $0.1467 \pm 0.0064$ & $0.1344 \pm 0.0006$ \\
        ContraWR & 1.6M & $0.3357 \pm 0.0164$ & $0.0743$ & $0.1401 \pm 0.0008$\\
        CNN-Transformer  & 3.2M & $0.3224 \pm 0.0109$ & $0.0628 \pm 0.0089$ & $0.1411 \pm 0.0007$ \\
        FFCL  & 2.4M & $0.3158 \pm 0.0235$ & $0.712 \pm 0.0124$ & $0.1396 \pm 0.0014$ \\
        ST-Transformer & 3.5M & $0.5442 \pm 0.0012$ & $0.2911 \pm 0.0014$ & $0.12222 \pm 0.0001$ \\
        BIOT & 3.2M &  $0.2757 \pm 0.0173$ & $0.0597 \pm 0.0069$ & $0.1401 \pm 0.0006$\\
        LaBraM-Base & 5.8M & $0.5383 \pm 0.0102$ & $0.2876 \pm 0.0032$ & $0.1225 \pm 0.0003$ \\
        LaBraM-Large & 46M & $0.5603 \pm 0.0020$ & $0.3093 \pm 0.0032$ & $0.1197 \pm 0.0003$ \\
        LaBraM-Huge & 369M & $0.5632 \pm 0.0023$ & $0.3145 \pm 0.0032$ & $0.1196 \pm 0.0003$ \\
        \hline
        \hline
    \end{tabular}%
    }
\end{table}

\subsection{Seed-V Dataset}

Another dataset we evaluate on is the Seed-V dataset \cite{liu2021comparing}, an emotion EEG dataset with five emotion categories: happy, sad, neural, disgust, and fear. The experiment has 10 male and 10 female subjects and were again truncated to the common 19 channels and downsampled to 128 Hz. Each participant participated in three sessions and each session included fifteen video clips corresponding to the five emotions. Unlike LaBraM, we segment into $6$-second samples (LaBraM segments into 1-second samples) and do not further split the training into training and validation, resulting in 10 trials for training and 5 trials for validation. Like LaBraM \cite{LABRAM}, because the labels are well balanced, we consider accuracy rather than balanced accuracy. The results are shown in Table \ref{tbl:SeedVResultTable}. Note that the other results from other models are from \cite{LABRAM}.

\begin{table}[ht]
    \centering
    \caption{Results on Seed-V Dataset}
    \label{tbl:SeedVResultTable}
    \resizebox{\linewidth}{!}{%
    \begin{tabular}{ccccc}
        \hline
        \hline
        & & \multicolumn{3}{c}{Seed-V (Multi-class Classification)}\\
        \cmidrule(r){3-5}
        Methods & Model Size & Accuracy & Cohen's Kappa & Weighted F1 \\
        \hline
        \textbf{MENDR-Tiny} & 1.815M &  $0.3839 \pm 0.0016$ & $0.2096 \pm 0.0016$ & $0.3996 \pm 0.0006$ \\
        \textbf{MENDR-Large} & 1.817M & $0.3748 \pm 0.0025$ & $0.2069 \pm 0.0018$ & $0.3854 \pm 0.0009$ \\
        \hline
        SPaRCNet  & 0.79M & $0.2887 \pm 0.0047$ & $0.1032 \pm 0.0083$ & $0.2904 \pm 0.0064$ \\
        ContraWR & 1.6M & $0.3603 \pm 0.0098$ & $0.1988 \pm 0.0114$ & $0.3590 \pm 0.0091$ \\
        CNN-Transformer  & 3.2M & $0.3665\pm 0.0058$ & $0.2034 \pm 0.0.0060$ & $0.3639 \pm 0.0065$ \\
        FFCL  & 2.4M & $0.3686 \pm 0.0059$ & $0.2094 \pm 0.0078$ & $0.3679 \pm 0.0062$ \\
        ST-Transformer & 3.5M & $0.2772 \pm 0.0047$ & $0.0783 \pm 0.0071$ & $0.3809 \pm 0.0114$ \\
        BIOT & 3.2M &  $0.3802 \pm 0.0094$ & $0.2247 \pm 0.0100$ & $0.3809 \pm 0.0114$\\
        LaBraM-Base & 5.8M & $0.4095 \pm 0.0062$ & $0.2613 \pm 0.0075$ & $0.4120 \pm 0.0057$ \\
        LaBraM-Large & 46M & $0.4096 \pm 0.0075$ & $0.2639 \pm 0.0090$ & $0.4127 \pm 0.0079$ \\
        LaBraM-Huge & 369M & $0.4102 \pm 0.0037$ & $0.2646 \pm 0.0046$ & $0.4136 \pm 0.0047$ \\
        \hline
        \hline
    \end{tabular}%
    }
\end{table}


\end{document}